\documentclass{article}

\usepackage[nonatbib,preprint]{main_sty}
\usepackage[utf8]{inputenc} % allow utf-8 input
\usepackage[T1]{fontenc}    % use 8-bit T1 fonts
\usepackage{hyperref}       % hyperlinks
\usepackage{url}            % simple URL typesetting
\usepackage{booktabs}       % professional-quality tables
\usepackage{amsfonts}       % blackboard math symbols
\usepackage{nicefrac}       % compact symbols for 1/2, etc.
\usepackage{microtype}      % microtypography
\usepackage{xcolor}         % colors

\RequirePackage{fancyhdr}
\RequirePackage{algorithm}
\RequirePackage{algorithmic}
\RequirePackage{eso-pic} % used by \AddToShipoutPicture
\RequirePackage{forloop}
\usepackage[numbers]{natbib}
\usepackage{amsmath}
\usepackage{amssymb}
\usepackage{mathtools}
\usepackage{amsthm}
\usepackage{cleveref}
\crefname{equation}{Eq.}{Eqs.}
\crefname{table}{Table}{Tables}
\crefname{figure}{Figure}{Figures}
\crefname{section}{Section}{Sections}
\crefname{algorithm}{Algorithm}{Algorithms}
\theoremstyle{plain}
\newtheorem{theorem}{Theorem}[section]
\newtheorem{proposition}[theorem]{Proposition}

\theoremstyle{definition}

\theoremstyle{remark}

\usepackage{xcolor,colortbl}
\usepackage{adjustbox}
\usepackage{url}
\usepackage{multirow,multicol,xspace}
\usepackage{float}
\usepackage{graphics}
\usepackage{paralist}
\usepackage{wrapfig}
\usepackage[normalem]{ulem}
\usepackage{pgfplots}
\pgfplotsset{width=8cm,compat=1.17} % <----
\usepackage{pifont}% http://ctan.org/pkg/pifont
\usepackage{subcaption,ragged2e}
\usepackage{caption}

\newcommand{\Zp}{\Tilde{Z}}

\newcommand{\x}{\mathcal{X}}

\newcommand{\chembl}{{ChEMBL2K}\xspace}
\newcommand{\broad}{{Broad6K}\xspace}
\newcommand{\toxcast}{ToxCast\xspace}
\newcommand{\biogen}{{Biogen3K}\xspace}

\newcommand{\method}{{InfoAlign}\xspace}
\definecolor{LightCyan}{rgb}{0.88,1,1}
\usepackage{soul}

\title{Learning Molecular Representation in a Cell}

\author{%
  Gang Liu$^1$, \quad Srijit Seal$^2$, \quad John Arevalo$^2$, \quad  Zhenwen Liang$^1$ \\
  \quad \textbf{Anne E. Carpenter}$^2$, \quad \textbf{Meng Jiang}$^1$, \quad \textbf{Shantanu Singh}$^2$ \\
  $^1$University of Notre Dame \quad $^2$Broad Institute of MIT and Harvard \\
  \texttt{\{gliu7, zliang6, mjiang2\}@nd.edu} \\
  \texttt{\{seal, jarevalo, anne, shantanu\}@broadinstitute.org} \\
}

\begin{document}

\maketitle

\begin{abstract}
Predicting drug efficacy and safety \textit{in vivo} requires information on biological responses (e.g., cell morphology and gene expression) to small molecule perturbations. However, current molecular representation learning methods do not provide a comprehensive view of cell states under these perturbations and struggle to remove noise, hindering model generalization.
We introduce the \textbf{Info}rmation \textbf{Align}ment (InfoAlign) approach to learn molecular representations through the information bottleneck method in cells. We integrate molecules and cellular response data as nodes into a context graph, connecting them with weighted edges based on chemical, biological, and computational criteria.
For each molecule in a training batch, InfoAlign optimizes the encoder's latent representation with a minimality objective to discard redundant structural information. 
A sufficiency objective decodes the representation to align with different feature spaces from the molecule's neighborhood in the context graph.
We demonstrate that the proposed sufficiency objective for alignment is tighter than existing encoder-based contrastive methods. 
Empirically, we validate representations from InfoAlign in two downstream applications: molecular property prediction against up to 27 baseline methods across four datasets, plus zero-shot molecule-morphology matching.
\end{abstract}

\section{Introduction}
\label{sec:introduction}
Drug properties, e.g., toxicity and adverse effects~\citep{liu2023using}, are induced by molecular initiating events—interactions between a molecule and a biological system—that first impact the cellular level and ultimately influence tissue or organ functions~\citep{mast2014systems}. 
However, a chemical molecule's structure alone is insufficient information to predict its impact on cells: each chemical interacts with multiple cells and genes and induces complex changes in gene expression and cell morphology, making predictions of downstream responses challenging~\citep{carpenter2006cellprofiler,moshkov2023predicting}. 
Hence, \textit{molecular representation learning should make use of information about cellular response}, enhancing the representation of the mode of action and thereby improving predictions for downstream bioactivity tasks~\citep{liu2023using,wang2023removing}.

There is a lack of exploration for holistic molecular representations from molecular structure, cell morphology, and gene expression \citep{hu2020strategies, you2020graph, liu2022graph,wang2023removing,sanchez2023cloome}. For example, graph self-supervised methods only manipulate molecular structures to perturb or mask molecular graphs using contrastive or predictive losses~\citep{hu2020strategies, you2020graph, inae2023motif}. \citet{moshkov2023predicting} explored the ability of different data modalities, taken independently, to predict molecules' assay activity in a diverse set of assays (tasks). They found (from \citep{moshkov2023predicting}'s Fig.2) that molecular structure supports highly accurate prediction (AUC > 90\%) in 31\% (16/52) of tasks, gene expression in 37\% (19/52) and cell morphology in 54\% (28/52).
Similarly, in our experiments (\cref{fig:target-bar}), we observe that molecular structure is not a one-size-fits-all solution.
\begin{figure*}[t]
    \centering
    \includegraphics[width=0.95\textwidth]{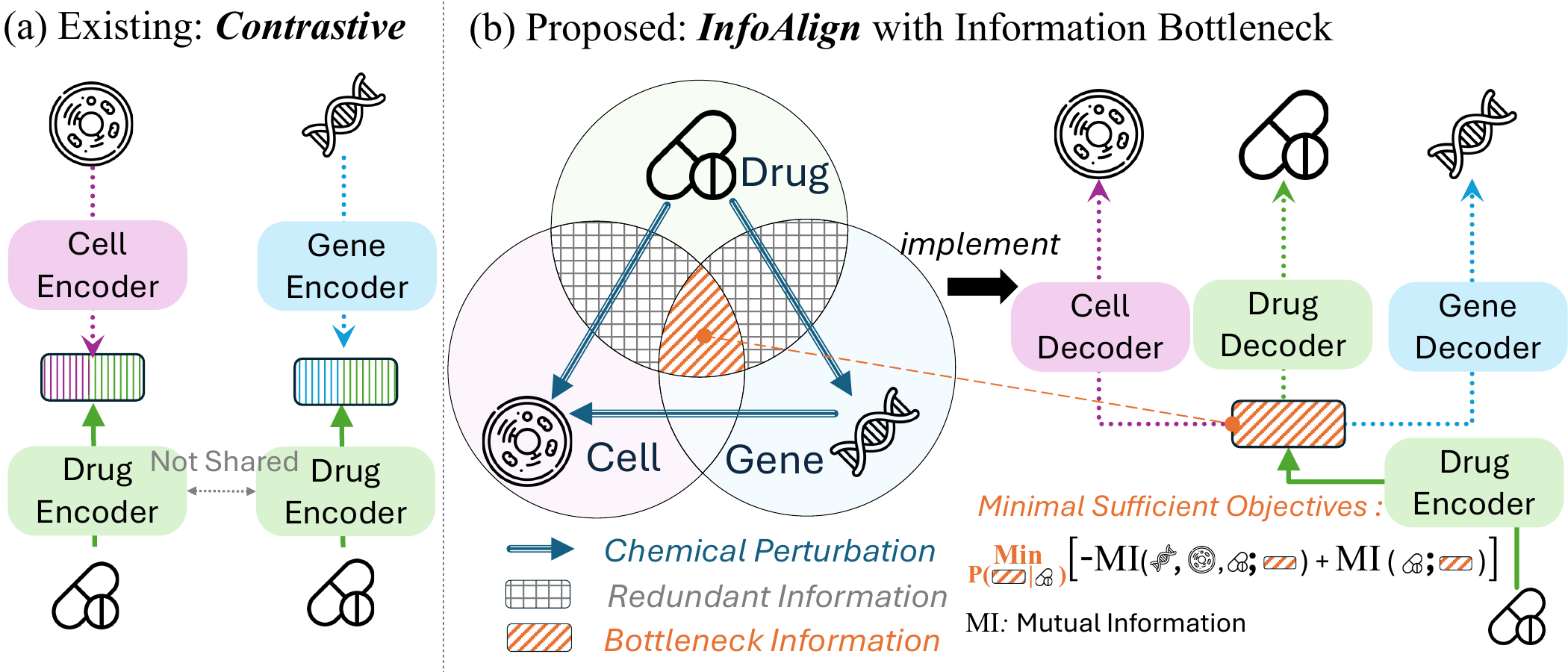}
    \caption{Comparison of Representation Learning Methods: (a) Existing contrastive methods use two encoders—one for molecules and another for cell morphology or gene expression features—without sharing the molecule encoders for different alignment targets.
    (b) \method remove redundant information from molecules, cell morphology, and gene expressions based on the information bottleneck, resulting in more concise yet predictive molecular representations~\citep{alemi2016deep}.}
    \label{fig:idea}
\end{figure*}

Cells can be perturbed by treating them with chemicals or genetic reagents that disrupt a particular gene or pathway. 
These chemical and genetic perturbations \textit{in vitro} naturally bridge molecules with cell morphology and gene expression, as illustrated in~\cref{fig:idea} (b). 
However, multi-modal contrastive methods such as CLOOME~\citep{sanchez2023cloome} and InfoCORE~\citep{wang2023removing}, depicted in~\cref{fig:idea} (a), focus primarily on aligning molecular representations with cell morphology~\citep{sanchez2023cloome,wang2023removing} or gene expression~\citep{wang2023removing}. 
These approaches fall short in two ways. 
(1) They do not remove redundant information, grey-colored area in~\cref{fig:idea} (b), that may harm representation generalization. 
The presence of redundant information~\citep{wang2023removing} may induce spurious correlations, adversely affecting the generalization of molecular representations. 
For example, in small molecule perturbations~\citep{bray2016cell,chandrasekaran2023jump}, batch identifiers can signify confounding technical factors, creating misleading associations between molecular structures and cell morphology~\citep{wang2023removing}.
(2) They treat molecules as the sole connectors between gene expression and cell morphology, ignoring the potential for genetic perturbations~\citep{chandrasekaran2023jump} to directly establish connections between these modalities. 
Genetic perturbations~\citep{chandrasekaran2023jump} not only enrich the feature space of gene expression and cell morphology but also enhance the navigation of molecular representation learning towards the overlapped (bottleneck) area in~\cref{fig:idea} (b).

To address the aforementioned challenges, we conceptualize the cellular response processes as a context graph, capturing a more complete set of interactions among molecules, gene expression, and cell morphology.
We identify the neighborhood of the molecule on the context graph and apply the information bottleneck~\citep{tishby2000information} to optimize molecular representations, which aligns them with neighboring biological variables to remove redundant information and improve generalization.

We propose the \textbf{Info}rmation \textbf{Align}ment (\method) approach, as presented in \cref{fig:idea} (b). \method uses one encoder and multiple decoders with information bottleneck for minimal sufficient statistics in representation learning. 
The minimality objective optimizes the encoder to learn the \textit{minimal} informative representation from molecular structures by discarding redundant information.
The sufficiency objective ensures the encoder retains \textit{sufficient} information, allowing decoders to reconstruct features for biological variables in neighborhood areas of the context graph.
We construct the context graph based on molecule and genetic perturbations~\citep{bray2017dataset,chandrasekaran2023jump,subramanian2017next} and introduce more biological (gene-gene interaction~\citep{himmelstein2017systematic}) and computational (cosine similarity) criteria to increase edge connectivity. 
We conduct random walks on the context graph, beginning with the molecule in the training batch, to identify its neighborhood. Cumulative edge weights indicate similarity between the molecule and variables along the path. 
The molecule is encoded, and its latent representation is decoded to align with features identified in the random walk. 
Encoders and decoders are jointly optimized using an upper bound for the minimality objective and a lower bound for the sufficiency objective.

The sufficiency objective introduces a decoder-based bound for multi-modal alignment. We show its theoretical advantages by demonstrating that it provides a tighter bound than the encoder-based approaches used in previous contrastive methods~\citep{oord2018representation,radford2021learning}, as discussed in~\cref{subsec:theory}. In experiments, \method outperforms up to \textbf{27} baselines across three classification and one regression dataset, covering 685 tasks, with average improvements of up to 6.4\%. \method also demonstrates strong zero-shot multi-modal matching on two molecule-morphology datasets.

\section{Related Work}
\label{sec:related}
\textbf{Representation Learning on Molecular Structure:}
Representation learning approaches for molecules can be categorized into sequential-based~\citep{krenn2022selfies,ross2022large} or graph-based models~\citep{hu2020strategies,you2020graph,zhang2021motif,liu2023semi}.
Sequential models, utilizing string formats of molecules like SMILES and SELFIES~\citep{krenn2022selfies}, have evolved from Recurrent Neural Networks (RNNs) to Transformers~\citep{chithrananda2020chemberta,ross2022large}. These models typically follow specific pretraining strategies similar to language models such as BERT~\citep{devlin2018bert}, RoBERTa~\citep{liu2019roberta,chithrananda2020chemberta} and GPT~\citep{radford2019language}. The pretraining targets are thus often the next token predictions or mask language modeling~\citep{devlin2018bert,chithrananda2020chemberta} on SMILES or SELFIES sequences~\citep{radford2019language}.
Graph Neural Networks (GNNs) are the architectures for graph-based approaches~\citep{hu2020strategies,you2020graph,zhang2021motif,liu2024graph}, where methods to pretrain GNNs often perturb or mask the atoms, edges, or substructures of molecular graphs with contrastive~\citep{hu2020strategies,you2020graph} and predictive losses~\citep{zhang2021motif,inae2023motif}. 
Recent evidence highlights the challenges of developing universal molecular representations based solely on molecular structures without integrating domain knowledge~\citep{bray2016cell,seal2022integrating,sun2022does,seal2023merging,liu2024data}. Although using motifs is a common method to incorporate such knowledge~\citep{rong2020self,inae2023motif}, the incorporation of information about molecules' biological impacts is much less explored. We aim to enhance molecular representation learning by incorporating domain knowledge from cellular response data.

\textbf{Representation Learning with Different Modalities:}
Existing methods on multimodal alignment, such as CLIP~\citep{radford2021learning}, primarily address pairwise relationships between texts and images and use methods like InfoNCE~\citep{oord2018representation,wang2023removing,sanchez2023cloome}. These approaches use separate encoders for different modalities to compute contrastive loss, which is upper bounded by the number of negative examples~\citep{poole2019variational}. Subsequent research on molecules similarly focuses on pairwise alignment between molecules and cell images~\citep{sanchez2023cloome,wang2023removing}, molecules and protein sequences~\citep{huang2021moltrans}, and molecules and text~\citep{edwards2022translation,jin2023large}. Although BioBridge~\citep{wang2023biobridge} handles multiple modalities, it leverages a knowledge graph for transforming representations between modalities rather than optimizing molecular representations.

\textbf{Representation Learning with Cellular Response Data:}
A primary goal of molecular representation learning is to predict molecular bioactivity. Likewise, emerging gene expression~\cite{subramanian2017next} and morphological profiling approaches~\cite{carpenter2006cellprofiler,seal2024decade} that describe perturbed genetic or cellular states in cell cultures can also be used to predict bioactivity. In some datasets, molecules are the perturbations, and the perturbed cell states measured are gene expression values for a thousand or more genes~\citep{subramanian2017next} and/or microscopy Cell Painting images, which can be represented as a thousand or more morphology features~\cite{cimini2023optimizing}. 
Recently created large-scale perturbation datasets~\citep{bray2016cell,subramanian2017next,chandrasekaran2023jump} could enrich molecular representation learning approaches. CLOOME~\citep{sanchez2023cloome} and MoCoP~\citep{nguyen2023molecule} contrast cellular images with molecules and InfoCORE~\citep{wang2023removing} contrasts molecule with either morphological profiling~\citep{bray2017dataset} or gene expression~\citep{wang2023removing}.  InfoCORE~\cite{wang2023removing} aims to mitigate confounding batch identifiers, but its effectiveness depends on a batch classifier, which is impractical without batch identifiers.
We integrate cellular response data and molecules into a context graph to capture cellular response patterns, focusing on learning molecular representations to remove nuisances~\citep{tian2020makes}.

\section{Problem Definition}
\label{sec:problem}
We denote $x \in \x$ as the molecule from the space $\x$. An encoder with parameters $p_\theta(\mathbf{z} \mid x)$  maps $x$ to a $D$-dimensional latent representation $\mathbf{z} \in \mathbb{R}^D$. One may implement a Graph Neural Network (GNN)~\citep{xu2018powerful} as the encoder. The GNN first updates node representations and then performs a readout operation (e.g., summation) over the nodes to obtain the latent representation.

Existing research has extensively used structural features to pretrain the GNN encoder~\citep{hu2020strategies,inae2023motif}. However, incorporating more expressive features from the cellular context, such as cell morphology and gene expression, remains largely unexplored for improving molecular representations. In this work, we use these features as targets to optimize molecular representations.

\section{Multi-modal Alignment with \method}
\label{sec:method}
\begin{figure*}[t]
    \centering
    \includegraphics[width=1\textwidth]{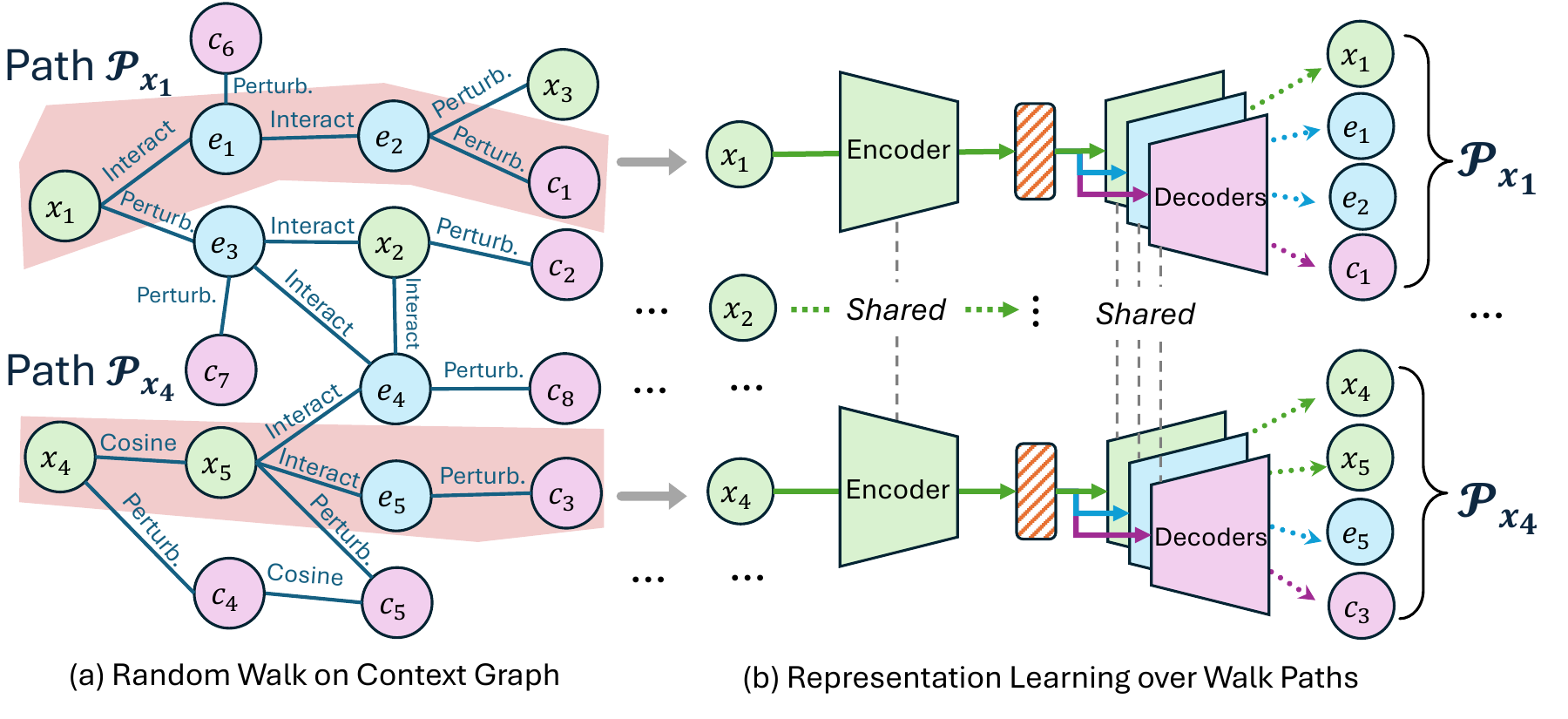}
    \caption{Molecular Representation Learning Using the Context Graph: (a) In \cref{subsec:context-graph-walk}, we construct the graph with various interaction, perturbation, and cosine similarities among molecules $x$, cell morphology profiles $c$, and genes $e$. Given a training batch of molecules, including $x_1$ and $x_4$, random walk extracts paths, for instance, of length four. (b) In \cref{subsec:represent-learn}, we aim to learn molecular representations based on the information bottleneck, preserving minimal information from the input molecule while ensuring sufficient information for decoding the target along the walk path $\mathcal{P}_x$.}
    \label{fig:framework}
    \vspace{-0.2in}
\end{figure*}

We present the overall representation learning framework in \cref{fig:framework}. In \cref{subsec:context-graph-walk}, we construct the context graph for cellular response data. 
In \cref{subsec:represent-learn}, we introduce representation learning methods based on the principle of minimal sufficiency for molecules and their related modalities. In \cref{subsec:theory}, we demonstrate the theoretical advantages of the proposed method.

\subsection{Random Walks on Cellular Context Graph}\label{subsec:context-graph-walk}

\textbf{Node Construction:}
We model the interactions of the molecule $x$ with other molecules, the cell $c$, and genes $e$ using the context graph. They are nodes with different features $y$.
Molecular features are vectors obtained using fingerprint~\citep{rogers2010extended}. Cell morphology features are vectors derived from CellProfiler \citep{carpenter2006cellprofiler} applied to Cell Painting microscopy images. Gene expression features are scalars using L1000 \citep{subramanian2017next} methods. We further rescale the feature spaces to a range between 0 and 1. 

\textbf{Edge Construction:}
We link nodes using various chemical, biological, and computational criteria. For example, molecules can perturb cultured human cells, inducing changes in cell morphology~\citep{chandrasekaran2023jump} and gene expression \citep{subramanian2017next}, thus linking them to cell morphology and gene expression nodes. Genes could also perturb cells, inducing links between genes and cell morphology~\citep{chandrasekaran2023jump}.
Additionally, we calculate cosine similarity within the same feature space and use biological criteria such as gene-gene interactions~\citep{himmelstein2017systematic} to enrich the edge space. Each edge is assigned a weight $w$ with a value between 0 and 1. We construct the context graph with details~\cref{sec:pretrain-setup}. An example is provided in~\cref{fig:framework} (a).

\textbf{Random Walk Path Extraction:}
The context graph identifies related cellular response patterns for input molecules in representation learning. Given an input molecule $x$, we extract its neighborhood through random walks starting from $x$. Specifically, we employ degree-based transition probabilities~\citep{perozzi2014deepwalk} and denote the walking path as $\mathcal{P}_x: x \xrightarrow[]{w_1} v_2 \xrightarrow[]{w_2} \dots \xrightarrow[]{w_L} v_L$, where $v_2$ is a direct neighbor of $x$. To quantify the similarity between $x$ and node $v_i$ ($2 \leq i \leq L$) on $\mathcal{P}_x$, we compute the cumulative product of edge weights as $\alpha(v_i \mid \mathcal{P}_x) = \prod_{j=1}^{i-1} w_j$.

\subsection{Optimization for Representation with Information Bottleneck}\label{subsec:represent-learn}

The information bottleneck (IB)~\citep{tishby2000information,alemi2016deep} is an appealing method for defining concise representations with strong predictive power. For molecular representation, we extract minimal sufficient information from the random variable $X$ of molecules. This is achieved by aligning the molecular representations $Z$ with the targets $Y$, derived from node features along the walk path $\mathcal{P}$. The IB has two principles based on mutual information (MI): (1) the minimality principle, which minimizes MI between molecules and their latent representations as $I(X; Z)$, and (2) the sufficiency principle, which decodes latent representations to maximally reconstruct feature spaces for variables along the walk path $ I \left(Z; Y \right)$. Together, these form the optimization objectives:
\begin{equation}\label{eq:mi-original-min}
    \min_{p(\mathbf{z} \mid x)}  \left[ - I \left(Z; Y \right) + \beta I(X; Z) \right],
\end{equation}
where $\beta$ controls the trade-off between minimality and sufficiency. The exact computation of $I(Z; Y)$ and $I(X; Z)$ is intractable due to the unknown conditional distribution $p(y | \mathbf{z})$ and the marginal $p(\mathbf{z})$. We introduce the variational approximations $q(y | \mathbf{z})$ and $q(\mathbf{z})$ for them, respectively. This results in a lower bound estimation for the first decoding term $I_{DLB}$ and an upper bound for the second encoding term $I_{EUB}$~\citep{poole2019variational}.
\begin{equation}
\begin{aligned}
    I(Z; Y) &\geq \mathbb{E}_{p(\mathbf{z}, y)} \left[ \log q(y \mid \mathbf{z})\right] + H(Y) \triangleq I_{DLB} \\
    I(X; Z) & \leq \mathbb{E}_{p(x)} \left[\operatorname{KL}\left( p(\mathbf{z} \mid x) \| q(\mathbf{z}) \right) \right] \triangleq I_{EUB} \label{eq:mi-lower-upper-bo}
\end{aligned}
\end{equation}
$H(Y)$ is the differential entropy. Proofs are in~\cref{sec:proof-lower-upper}. Together, $I_{DLB}$ and $I_{EUB}$ upper bound \cref{eq:mi-original-min}, forming a tractable objective $-I_{DLB} + I_{EUB}$ to optimize the encoder.
For the target $Y$, the $I_{DLB}$ objective requires decoders rather than encoders, as typically used in prior work~\citep{sanchez2023cloome}. We use distinct decoders, denoted as $q_\phi$ with parameters $\phi$, for various targets, including molecular fingerprints, gene expressions, and cell morphology features.

After ignoring the constant terms, one could formulate the loss function according to~\cref{eq:mi-lower-upper-bo} for the molecule sample $x$, its latent representation $\mathbf{z}$, and the targets $y_v$ from $\mathcal{P}_x$:
\begin{equation}\label{eq:loss}
    \mathcal{L} = \frac{1}{L} \sum_{v \in \mathcal{P}_x} \alpha(v | \mathcal{P}_x) \left[ -\log \left(q_\phi (y_v \mid \mathbf{z} \right) \right] + \beta \operatorname{KL}\left(p_\theta(\mathbf{z} \mid x) \mid \mathcal{N}(0, I)\right),
\end{equation}
where the first term aligns the representation with other features, and $\operatorname{KL}$ is the Kullback–Leibler divergence used for regularization. $\mathcal{N}(0, I)$ is the The Gaussian prior. In this formulation, the encoder models a distribution instead of a single representation $\mathbf{z}$, learning the mean and variance $\boldsymbol{\mu}, \boldsymbol{\sigma} \in \mathbb{R}^D$. One may use parameterization tricks to sample $\mathbf{z}$ from the distribution~\citep{alemi2016deep}. The decoder then reconstructs $y_v$, the features of the neighboring node $v$ on the context graph.

\method uses multiple decoders for $q_\phi$ to align multi-modal features, while prior work relies on encoders with CLIP-like losses to align the latent space~\citep{radford2021learning,girdhar2023imagebind,wang2023removing,sanchez2023cloome}. Next, we provide the theoretical benefits of decoder-based alignment alongside the empirical advantages in~\cref{sec:experiments}.

\subsection{Theoretical Motivation for Decoder-based Alignment}\label{subsec:theory}

InfoNCE~\citep{oord2018representation} is the contrastive loss used for most CLIP-like methods~\citep{radford2021learning,wang2023removing}. In this work, we show that the MI lower bound based on InfoAlign is tighter than that based on InfoNCE.

\begin{proposition}\label{propos:bound-diff}

For the molecular representation $Z$ and target $Y$ (from cell morphology, gene expressions, or molecular fingerprints), the encoder-based MI lower bound $I_{ELB}$ for InfoNCE can be derived by incorporating $K-1$ additional samples, denoted as $y_{2:K}$, to build the Monte Carlo estimate $m(\cdot)$ of the partition function~\citep{nguyen2010estimating,poole2019variational}:
\begin{equation}\label{eq:elb}
    I_{ELB} = 1 + \mathbb{E}_{p(z, y)p(y_{2:K})} \left[ \log \frac{e^{h(z, y)}}{m(z; y, y_{2:K})} \right] - \mathbb{E}_{p(z)p(y_{2:K})p(y)} \left[ \frac{e^{h(z, y)}}{m(z; y, y_{2:K})} \right],
\end{equation}
where $h(z, y)$ is the neural network parameterized critic for density approximation with the energy-based variational family. The decoder-based lower bound $I_{DLB}$ is defined in~\cref{eq:mi-lower-upper-bo}, then we have that $I_{DLB}$ is tighter than $I_{ELB}$, i.e., $I(Z; Y) \geq I_{DLB}(Z; Y) \geq I_{ELB}(Z; Y)$.
\end{proposition}
Proofs are in~\cref{sec:proof-bound-diff}. The result aligns with empirical observations in previous studies such as DALL-E 2~\citep{ramesh2022hierarchical},  where a prior model was introduced to improve representations from CLIP~\citep{radford2021learning} before decoding to another modality. In this work, we learn decodable latent representations from molecules to align with different biological features.

\section{Implementation of Context Graph and Pretraining Setting}
\label{sec:pretrain-setup}
\textbf{Data Source of Context Graph:}
We create the context graph based on (1) two Cell Painting datasets~\citep{bray2017dataset,chandrasekaran2023jump}, containing around 140K molecule perturbations (molecule and cell morphology pairs) and 15K genetic perturbations (gene and cell morphology pairs) across 1.6 billion human cells; (2) Hetionet~\citep{himmelstein2017systematic}, which captures gene-gene and gene-molecule relationships from millions of biomedical studies; and (3) a dataset reporting differential gene expression values for 978 landmark genes~\citep{wang2016drug} for chemical perturbations (molecule and gene expression pairs)~\citep{subramanian2017next}.

\textbf{Node Features:} 
Different profiling methods provide node features in different ways. Morgan fingerprints \citep{rogers2010extended} are feature vectors extracted from each molecule's structure, CellProfiler features~\citep{carpenter2006cellprofiler} are computed from the image of each cell and represent cell morphology, and L1000 profiles~\citep{subramanian2017next} capture gene expression values on 978 landmark genes from cells treated with a chemical perturbation. Here are two practical considerations for the context graphs:
(1) ~\citet{chandrasekaran2023jump} provided one dataset that measured the cell morphology impacts of perturbing individual genes. The 15K genetic perturbations~\citep{chandrasekaran2023jump} provide gene-cell morphology pairs but lack corresponding gene expression profiles. Still, we keep the gene nodes from this dataset to account for potential gene-gene interactions and incorporate cell morphology features into them.
(2) All 978 landmark genes have expression values linked to the molecules used in~\citep{wang2016drug}. We update new gene expression nodes with 978-dimensional feature vectors. These vectors summarize all molecule-gene expression connections for a small molecule perturbation. This approach efficiently reduces dense connections between landmark genes and molecules. We select the top 1\% of gene-molecule expression values as new edges to enrich the context graph's connectivity.
We scale cell morphology and gene expression features to a range of 0 to 1 using the Min-Max scaler along each dimension.

\textbf{Edge Weights:}
For edges based on chemical perturbations~\citep{bray2017dataset,chandrasekaran2023jump}, we assign the edge weight of 1.
We also compute cosine similarity for nodes if they are in the same feature space (such as two cell morphology/gene expression profiles, or Morgan fingerprints). To avoid noisy edges from computations, we apply a 0.8 threshold for cosine similarity, and additionally explicitly enforce 99.5\% sparsity by selecting top similar edges.

All together, this results in a context graph of 276,855 nodes (129,592 molecules, 4533 genes + 13,795 gene expressions, and 128,935 cell morphology) and 366,384 edges.

\textbf{Encoder and Decoder:}
We use the Graph Isomorphism Model (GIN)~\citep{xu2018powerful} as the molecule encoder. All molecules on the context graph are used to pretrain the encoder.
Because we extract feature vectors as the decoding targets in different modalities, we could efficiently use Multi-Layer Perception (MLP) as modality decoders.
In each training batch, random walks start from the molecule node to extract the walk path. Then, decoders are pretrained to reconstruct corresponding node features from nodes over the path. 
More details are in~\cref{sec:more-context-graph}.

\section{Experiments}
\label{sec:experiments}
\begin{table*}[t]
    \caption{Datasets and task information. Classf. denotes classification and Regr. denotes regression.}
    \label{tab:task-data-info}
    \centering
    \begin{adjustbox}{width=1\linewidth}
    \begin{tabular}{llrrrrrr}
    \toprule
    \multirow{2}{*}{Dataset} & \multirow{2}{*}{Type} & \multirow{2}{*}{\# Task} & \multirow{2}{*}{\# Molecules}& \multirow{1}{*}{\# Atoms} & \multirow{1}{*}{\# Edges} & \# Available Cell & \# Available Gene \\
     &  &  &  & Avg./Max & Avg./Max & Morphology & Expressions \\
    \midrule
    \chembl & Classf. & 41 & 2355 & 23.7/61 & 25.6/68 & 2353 & 631 \\
    \broad & Classf. & 32 & 6567 & 34.1/74 & 36.8/82 & 2673 & 1138 \\
    \toxcast & Classf. & 617 & 8576 & 18.8/124 & 19.3/134 & N.A. & N.A. \\
    \biogen & Regr. & 6 & 3521 & 23.2/78 & 25.3/84 & N.A. & N.A. \\
    \bottomrule
    \end{tabular}
    \end{adjustbox}
    % \vspace{-0.1in}
\end{table*}

We focus on three research questions (RQs) regarding \method's representation for molecular property prediction, molecule-morphology matching, and performance analysis.

\begin{table*}[t]
    \caption{Results on \chembl~and~\broad. We report average AUC (Avg.), as well as the percentage of tasks achieving AUC above 80\%, 85\%, and 90\%. We highlight the \textbf{best} and \underline{second best} mean. We also highlight the \colorbox{LightCyan}{row of the best method} in each category.}
    \vspace{-0.05in}
    \label{tab:chembl-broad-ext}
    \renewcommand{\arraystretch}{1.1}
    \renewcommand{\tabcolsep}{4pt}
    \centering
    \begin{adjustbox}{width=1\linewidth}
    \begin{tabular}{l|cccc|cccc}
        \toprule
        % \rowcolor{gray!20}
        Dataset & \multicolumn{4}{c|}{ChEMBL2k~(AUC $\uparrow$)} & 
        \multicolumn{4}{c}{Broad6k~(AUC $\uparrow$)} \\
        (\# Molecule / \# Task) & \multicolumn{4}{c|}{(2355 / 41)} & \multicolumn{4}{c}{(6567 / 32)} \\
        % \rowcolor{gray!20}
        Method & Avg. & >80\% & >85\% & >90\% & Avg. & >80\% & >85\% & >90\% \\
        \midrule
        % \rowcolor{gray!10}
        \multicolumn{9}{c}{\textit{Morgan Fingerprints}} \\
        % \multicolumn{1}{l}{\textbf{Best Morgan FP (MLP)}} & \multicolumn{1}{l}{76.8±2.2} & 48.8±3.9 & 34.6±6.3 & \multicolumn{1}{l}{21.9±5.7} & 63.3±0.3 & 6.3±0.0 & \textbf{4.4}±1.7 & \textbf{3.1}±0.0 \\
        \rowcolor{LightCyan}
        MLP & 76.8±2.2 & 48.8±3.9 & 34.6±6.3 & 21.9±5.7 & 63.3±0.3 & 6.3±0.0 & \textbf{4.4}±1.7 & \textbf{3.1}±0.0 \\
        RF  & 54.7±0.7 & 0.0±0.0 & 0.0±0.0 & 0.0±0.0 & 55.5±0.1 & 0.0±0.0 & 0.0±0.0 & 0.0±0.0 \\
        GP  & 51.0±0.0 & 0.0±0.0 & 0.0±0.0 & 0.0±0.0 & 50.6±0.0 & 0.0±0.0 & 0.0±0.0 & 0.0±0.0 \\
        \midrule
        % \rowcolor{gray!10}
        \multicolumn{9}{c}{\textit{Pretrained GNN}} \\
        AttrMask~\citep{hu2020strategies} & 73.9±0.5 & 46.8±2.7 & 31.2±4.4 & 14.6±1.7 & 59.8±0.2 & 3.1±0.0 & 3.1±0.0 & \textbf{3.1}±0.0 \\
        % \rowcolor{LightCyan}
        \cellcolor{LightCyan}ContextPred~\citep{hu2020strategies} & \cellcolor{LightCyan}77.0±0.5 & \cellcolor{LightCyan}55.1±1.3 & \cellcolor{LightCyan}34.1±4.6 & \cellcolor{LightCyan}14.6±1.7 & 60.0±0.2 & 7.5±1.7 & 3.1±0.0 & \textbf{3.1}±0.0 \\
        EdgePred~\citep{hu2020strategies}    & 75.6±0.5 & {54.2}±4.0 & 34.6±7.2 & 12.2±2.4 & 59.9±0.2 & 3.1±0.0 & 3.1±0.0 & \textbf{3.1}±0.0 \\
        GraphCL~\citep{you2020graph} & {75.6}±1.6 & 46.8±7.6 & 32.2±6.8 & 18.0±3.7 & 67.2±0.5 & \underline{15.6}±3.1 & 3.1±0.0 & \textbf{3.1}±0.0 \\
        GROVER~\citep{rong2020self} & 73.3±1.4 & 38.5±2.0 & 22.4±3.6 & 14.6±2.4 & 66.2±0.1 & \underline{15.6}±0.0 & \underline{3.8}±1.4 & \textbf{3.1}±0.0 \\
        \cellcolor{LightCyan}JOAO~\citep{you2020graph} & 75.1±1.0 & 47.8±5.1 & 33.7±2.0 & 19.0±3.2 & \cellcolor{LightCyan}\underline{67.3}±0.4 & \cellcolor{LightCyan}12.5±0.0 & \cellcolor{LightCyan}\underline{3.8}±1.4 & \cellcolor{LightCyan}\textbf{3.1}±0.0 \\
        MGSSL~\citep{zhang2021motif} & 75.1±1.1 & 39.0±4.6 & 29.3±3.0 & 10.3±3.2 & 66.9±0.5 & 13.8±2.8 & 3.1±0.0 & \textbf{3.1}±0.0 \\
        GraphLoG~\citeauthor{xu2021self}    & 73.5±0.7 & 41.9±2.0 & 29.3±3.4 & 15.6±2.8 & 62.9±0.4 & 4.4±1.7 & 0.0±0.0 & 0.0±0.0 \\
        GraphMAE~\citep{hou2022graphmae} & 74.7±0.1 & 33.2±1.3 & 27.8±1.3 & 12.2±1.7 & 66.8±0.3 & 14.4±1.7 & 3.1±0.0 & \textbf{3.1}±0.0 \\
        DSLA~\citep{kim2022graph}  & 69.3±1.0 & 23.9±4.7 & 14.6±5.5 & 6.8±1.1  & 63.3±0.3 & 6.3±0.0 & 3.1±0.0 & \textbf{3.1}±0.0 \\
        UniMol~\citep{zhou2023unimol} & 76.8±0.4 & 46.8±2.0 & 33.7±1.1 & 24.9±2.0 & 65.4±0.1 & 7.5±1.7 & 3.1±0.0 & \textbf{3.1}±0.0 \\
        \midrule
        % \rowcolor{gray!10}
        \multicolumn{9}{c}{\textit{Pretrained Chemical Language Models}} \\
        \cellcolor{LightCyan}Roberta~\citep{datamol2024} & \cellcolor{LightCyan}74.7±1.9 & \cellcolor{LightCyan}46.3±3.4 & \cellcolor{LightCyan}35.1±4.4 & \cellcolor{LightCyan}22.9±1.3 & 59.8±0.7 & 5.0±1.7 & 3.1±0.0 & \textbf{3.1}±0.0 \\
        \cellcolor{LightCyan}GPT2~\citep{datamol2024} & 71.0±3.4 & 31.2±11.2 & 20.0±9.4 & 7.3±6.9 & \cellcolor{LightCyan}60.6±0.3 & \cellcolor{LightCyan}7.5±1.7 & \cellcolor{LightCyan}1.9±1.7 & \cellcolor{LightCyan}1.9±1.7 \\
        MolT5~\citep{edwards2022translation}  & 69.9±0.8 & 32.2±2.0 & 21.0±4.1 & 8.8±1.3 & 56.4±0.8 & 3.8±1.4 & 2.5±1.4 & \underline{2.5}±1.4 \\
        ChemGPT~\citep{frey2023neural} & 65.0±1.1 & 16.1±2.8 & 11.2±3.3 & 5.4±1.1 & 55.1±0.9 & 3.1±0.0 & 3.1±0.0 & 1.3±1.7 \\
        \midrule
        % \rowcolor{gray!10}
        \multicolumn{9}{c}{\textit{Cell Morphology}} \\
        \cellcolor{LightCyan}MLP & \cellcolor{LightCyan}64.3±2.4 & \cellcolor{LightCyan}15.6±6.6 & \cellcolor{LightCyan}8.3±3.7 & \cellcolor{LightCyan}4.9±3.9 & 51.9±1.0 & 0.0±0.0 & 0.0±0.0 & 0.0±0.0 \\
        \cellcolor{LightCyan}RF  & 55.9±0.7 & 3.9±1.3  & 3.9±1.3  & 2.4±0.0  & \cellcolor{LightCyan}55.3±0.1 & \cellcolor{LightCyan}0.0±0.0  & \cellcolor{LightCyan}0.0±0.0  & \cellcolor{LightCyan}0.0±0.0  \\
        GP  & 50.1±0.0 & 0.0±0.0  & 0.0±0.0 & 0.0±0.0 & 54.7±0.0 & 0.0±0.0  & 0.0±0.0  & 0.0±0.0  \\
        \midrule
        % \rowcolor{gray!10}
        \multicolumn{9}{c}{\textit{Gene Expression}} \\
        \rowcolor{LightCyan}
        MLP & 56.1±1.1 & 5.1±1.4 & 3.4±1.3 & 3.4±1.3 & 56.9±1.4 & 1.9±1.7 & 1.9±1.7 & 1.9±1.7 \\
        RF  & 52.8±0.3 & 0.0±0.0 & 0.0±0.0 & 0.0±0.0 & 55.2±0.2 & 0.0±0.0 & 0.0±0.0 & 0.0±0.0 \\
        GP  & \multicolumn{4}{c|}{Run out of time} & 50.1±0.0 & 0.0±0.0 & 0.0±0.0 & 0.0±0.0 \\
        \midrule
        % \rowcolor{gray!10}
        \multicolumn{9}{c}{\textit{Multi-modal Alignment}} \\
        CLOOME & 66.7±1.8 & 26.8±4.6 & 16.1±3.7 & 10.7±5.1 & 61.7±0.4 & 3.1±0.0 & 3.1±0.0 & 0.0±0.0 \\
        InfoCore (GE) & \underline{79.3}±0.9 & \underline{62.4}±2.8 & \underline{46.3}±3.0 & \underline{30.3}±2.2 & 60.2±0.2 & 3.1±0.0 & 0.0±0.0 & 0.0±0.0 \\
        InfoCore (CP) & 73.8±2.0 & 37.6±9.2 & 26.3±4.7 & 10.7±4.1 & 61.1±0.2 & 6.3±0.0 & 3.1±0.0 & 0.0±0.0 \\
        \rowcolor{LightCyan}
        InfoAlign (Ours) & \textbf{81.3}±0.6 & \textbf{66.3}±2.7 & \textbf{49.3}±2.7 & \textbf{35.1}±3.7 & \textbf{70.0}±0.1 & \textbf{18.8}±2.2 & 3.1±0.0 & \textbf{3.1}±0.0 \\
        \bottomrule
    \end{tabular}
    \end{adjustbox}
    \vspace{-0.2in}
\end{table*}
\subsection{RQ1: Molecular Property Prediction}
\subsubsection{Experimental Setting}
\textbf{Dataset and Evaluation:}
We select datasets for important tasks including activity classification for various assays in \chembl~\citep{gaulton2012chembl} and \broad~\citep{moshkov2023predicting}, drug toxicity classification using \toxcast~\citep{richard2016toxcast}, and absorption, distribution, metabolism, and excretion (ADME) regression using \biogen~\citep{fang2023prospective}. The dataset statistics are in~\cref{tab:task-data-info}, covering \textbf{685} tasks. We apply scaffold-splitting for all datasets. We follow \citep{hu2020open} for the ToxCast dataset, and a 0.6:0.15:0.25 ratio for training, validation, and test sets for other datasets.
We use the Area under the curve (AUC) for classification and mean absolute error (MAE) for regression.
Mean and standard deviations are reported from ten runs.

\textbf{Baseline:}
We include \textbf{27} baselines across six categories: (1) three molecular fingerprint (FP)-based methods~\citep{rogers2010extended}; (2) eleven pretrained GNNs; (3) four pretrained chemical language models; (4,5) six methods based on cell morphology and gene expression values from cells treated with each molecule; (6) CLOOME~\citep{sanchez2023cloome} and InfoCORE~\citep{wang2023removing} for multi-modal alignment using structure, morphology, and gene expression data. We use MLPs, Random Forests (RF), and Gaussian Processes (GP) for methods in categories (1,4,5). We fine-tune MLPs on various representation learning approaches for predicting molecular properties. Setting details and all results are in~\cref{sec:more-exp-setup,sec:more-assay-result}.

\vspace{-0.1in}
\begin{table*}[t]
    \caption{Results on \toxcast~and~\biogen. We report the average AUC and the percentage of AUC above 80\% on \toxcast, and regression MAE (scaled by × 100) for \biogen. We highlight the \textbf{best} and \underline{second best} mean. We also highlight the \colorbox{LightCyan}{row of the best method} in each category.}
    % \vspace{-0.1in}
    \label{tab:tox-biogen-ext}
    \renewcommand{\arraystretch}{0.9}
    \renewcommand{\tabcolsep}{0.8mm}
    \centering
    \begin{adjustbox}{width=1.01\linewidth}
    \begin{tabular}{l|ll|lllllll}
    \toprule
    Dataset & \multicolumn{2}{c|}{\toxcast~(AUC $\uparrow$)} & \multicolumn{7}{c}{\biogen~(MAE $\times 100 \downarrow$ )} \\
    (\# Molecule / \# Task) & \multicolumn{2}{c|}{(8576 / 617)} & \multicolumn{7}{c}{(3521 / 6)} \\
    Method & Avg. & \textgreater 80 \% & Avg. & hPPB & rPPB & RLM & HLM & ER & Solubility \\
    \midrule
    % \rowcolor{gray!10}
    \multicolumn{10}{c}{\textit{Morgan Fingerprints}} \\
    \cellcolor{LightCyan}MLP & \cellcolor{LightCyan}57.6±1.0 & \cellcolor{LightCyan}1.6±0.3 & 66.2±2.4 & 66.1±2.6 & 56.8±2.3 & 56.5±4.2 & 74.6±6.2 & 73.7±7.3 & 69.5±3.0 \\
    \cellcolor{LightCyan}RF & 52.3±0.1 & 0.2±0.1 & \cellcolor{LightCyan}\underline{52.8}±0.2 & \cellcolor{LightCyan}44.2±0.1 & \cellcolor{LightCyan}\underline{44.2}±0.1 & \cellcolor{LightCyan}42.0±0.2 & \cellcolor{LightCyan}67.7±0.7 & \cellcolor{LightCyan}66.9±0.9 & \cellcolor{LightCyan}51.6±0.1 \\
    GP & \multicolumn{2}{c|}{Run out of Time} & 60.0±0.0 & 51.3±0.0 & 59.5±0.0 & 49.7±0.0 & 68.8±0.0 & 69.3±0.0 & 61.6±0.0 \\
    \midrule
    % \rowcolor{gray!10}
    \multicolumn{10}{c}{\textit{Pretrained GNN}} \\
    AttrMask~\citep{hu2020strategies} & 63.1±0.8 & 3.2±1.2 & 67.3±0.3 & 82.4±1.1 & 49.8±0.7 & 51.7±1.0 & \textbf{57.9}±0.6 & \underline{62.6}±0.5 & 99.1±1.2 \\
    ContextPred~\citep{hu2020strategies} & 63.0±0.6 & 3.3±1.3 & 68.5±0.9 & 85.0±7.9 & 49.7±0.4 & 55.1±2.7 & \underline{61.4}±1.8 & 63.1±0.5 & 96.5±3.7 \\
    EdgePred~\citep{hu2020strategies} & 63.5±1.1 & 4.8±3.0 & 67.8±0.9 & 81.2±10.2 & 48.0±0.5 & 53.5±2.8 & 62.2±1.8 & 62.9±0.7 & 99.1±6.9 \\
    GraphCL~\citep{you2020graph} & 52.2±0.2 & 0.5±0.3 & 53.9±0.6 & 43.8±0.3 & 45.4±0.6 & 40.6±0.5 & 76.7±1.0 & 67.1±2.2 & 49.6±0.3 \\
    GROVER~\citep{rong2020self} & 53.1±0.4 & 0.5±0.1 & 54.9±1.6 & 44.5±0.4 & 46.5±0.7 & 41.7±0.6 & 73.2±5.7 & 71.0±4.3 & 52.6±0.3 \\
    JOAO~\citep{you2021graph} & 52.3±0.2 & 0.4±0.1 & 55.0±0.8 & 44.5±0.5 & 47.6±0.5 & \underline{40.6}±0.2 & 74.3±2.8 & 71.5±2.6 & 51.4±0.6 \\
    MGSSL~\citep{zhang2021motif} & 64.2±0.2 & 4.0±0.4 & 53.2±0.3 & 44.8±0.6 & 49.7±0.3 & 41.5±0.2 & 65.6±1.8 & 64.6±0.5 & 52.7±0.5 \\
    GraphLoG~\citep{xu2021self} & 58.6±0.4 & 2.5±0.3 & 56.9±0.4 & 49.3±0.3 & 54.8±0.5 & 42.6±0.3 & 66.8±1.7 & 69.0±1.3 & 58.8±0.5 \\
    \cellcolor{LightCyan}GraphMAE~\citep{hou2022graphmae} & 53.3±0.1 & 0.6±0.1 & \cellcolor{LightCyan}\underline{52.8}±0.8 & \cellcolor{LightCyan}\underline{43.3}±0.9 & \cellcolor{LightCyan}51.2±0.8 & \cellcolor{LightCyan}40.9±0.3 & \cellcolor{LightCyan}64.4±2.7 & \cellcolor{LightCyan}65.9±3.8 & \cellcolor{LightCyan}\underline{50.9}±1.4 \\
    DSLA~\citep{kim2022graph} & 57.8±0.5 & 0.7±0.1 & 57.9±0.7 & 50.4±0.7 & 53.6±1.7 & 43.3±0.9 & 68.6±1.2 & 70.8±2.0 & 60.9±0.6 \\
    \cellcolor{LightCyan}UniMol~\citep{zhou2023unimol} & \cellcolor{LightCyan}64.6±0.2 & \cellcolor{LightCyan}4.8±1.0 & 55.8±2.8 & 50.1±5.2 & 49.9±5.6 & 43.6±1.1 & 65.4±4.9 & 65.8±1.2 & 59.9±6.6 \\
    \midrule
    % \rowcolor{gray!10}
    \multicolumn{10}{c}{\textit{Pretrained Chemical Language Models}} \\
    Roberta~\citep{datamol2024} & 64.2±0.8 & 3.1±1.8 & 69.0±2.6 & 71.4±14.5 & 65.1±19.2 & 63.7±24.6 & 67.5±5.2 & 69.9±4.9 & 76.7±13.2 \\
    GPT2~\citep{datamol2024} & 61.5±1.1 & 2.4±0.6 & 74.0±8.5 & 65.4±12.9 & 73.1±20.8 & 54.1±12.9 & 83.2±21.5 & 86.1±19.8 & 81.8±25.5 \\
    \rowcolor{LightCyan}
    MolT5~\citep{edwards2022translation} & 64.7±0.9 & 3.6±1.1 & 65.1±0.5 & 76.7±2.1 & 55.9±1.1 & 49.2±1.0 & 70.3±0.8 & 73.1±1.0 & 65.3±1.7 \\
    ChemGPT~\citep{frey2023neural} & \multicolumn{2}{c|}{Token Error} & 75.7±8.5 & 59.5±7.3 & 88.8±32.3 & 76.1±11.8 & 84.0±20.6 & 77.2±8.5 & 68.6±7.1 \\
    \midrule
    % \rowcolor{gray!10}
    \multicolumn{10}{c}{\textit{Multi-modal Alignment}} \\
    CLOOME & 54.2±0.9 & 0.9±0.2 & 64.3±0.4 & 65.2±1.5 & 56.9±0.8 & 44.2±0.8 & 70.7±0.4 & 73.6±0.8 & 75.0±2.1 \\
    InfoCORE (GE) & \underline{65.3}±0.2 & \underline{5.4}±1.7 & 69.9±1.2 & 79.9±3.6 & 51.6±1.8 & 51.3±2.1 & 78.6±0.3 & 77.8±1.9 & 80.3±0.9 \\
    InfoCORE (CP) & 62.4±0.4 & 1.3±0.5 & 71.0±0.6 & 74.5±4.9 & 53.5±0.7 & 53.6±2.1 & 80.8±1.5 & 79.4±3.4 & 84.4±1.0 \\
    \rowcolor{LightCyan}
    InfoAlign (Ours) & \textbf{66.4}±1.1 & 6.6±1.6 & \textbf{49.4}±0.2 & \textbf{39.7}±0.4 & \textbf{39.2}±0.3 & \textbf{40.5}±0.6 & 66.7±1.7 & \textbf{62.0}±1.5 & \textbf{48.4}±0.6 \\
    \bottomrule
    \end{tabular}
    \end{adjustbox}
    \vspace{-0.1in}
\end{table*} 
\subsubsection{Results and Analysis}

\begin{figure}[t]
    \vspace{-0.15in}
    \centering
    \begin{subfigure}[b]{0.48\textwidth}
        \centering
        \includegraphics[width=1\linewidth]{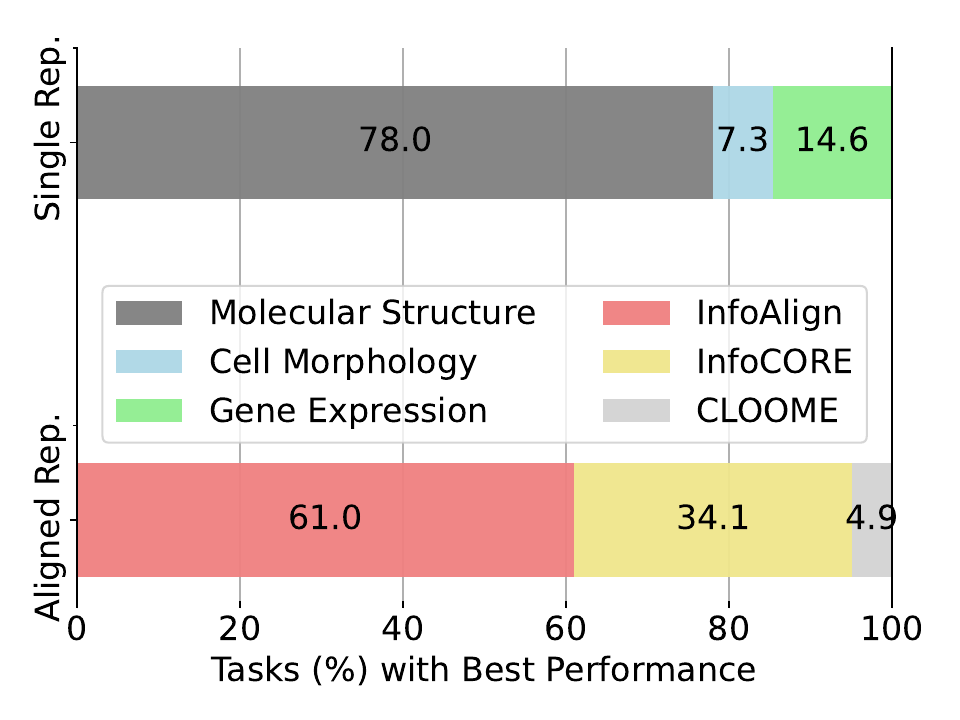}
        \caption{\chembl~with 41 Tasks.} % Optional; add if needed
        \label{fig:chembl-target-pie}
    \end{subfigure}%
    \hfill
    \begin{subfigure}[b]{0.48\textwidth}
        \centering
        \includegraphics[width=1\linewidth]{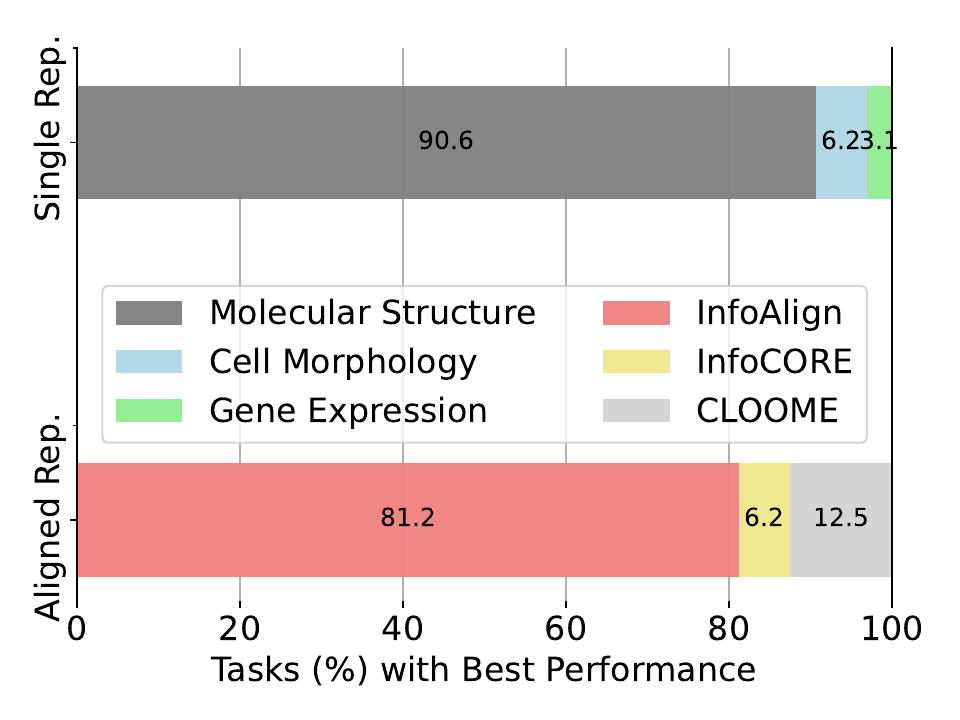}
        \caption{\broad~with 32 Tasks.} % Optional; add if needed
        \label{fig:broad-target-pie}
    \end{subfigure}
    \caption{Percentage of Tasks Where Representations Excel: We compare the relative performance of three single representation (Single Rep.) approaches (molecular structure, cell morphology, and gene expression) and three aligned representations (Aligned Rep.): \method, CLOOME, InfoCORE.}
    \label{fig:target-bar}
    % \vspace{-0.2in}
\end{figure}

We present results across various assays in~\cref{tab:chembl-broad-ext,tab:tox-biogen-ext} and~\cref{fig:target-bar}. Key observations include:

(1) \textbf{Molecular structures are superior compared to cell morphology and gene expression features for predicting various molecular assays}. This is likely because the datasets and tasks we selected fundamentally involve predicting the binding affinity of a molecule to a protein~\cite{gaulton2012chembl}; furthermore, in these datasets, molecules with activity in a given assay tend to have highly related structures, rather than representing two or more structurally distinct classes of molecules with activity; together this implies that molecular structure alone will tend to yield strong results. When comparing the three popular structure-based representation approaches, no single method outperforms the others across all four datasets. Pretrained GNNs generally perform better than fingerprint-based methods and pretrained chemical language models, thanks to recent advancements. However, continued efforts in universal structural representation are still necessary.

(2) \textbf{Cell morphology and gene expression features may complement molecular structures, yielding more generalizable representations}. 
As shown in \cref{fig:target-bar}, cell morphology and gene expression outperform molecular structure in approximately 20\% and 10\% of tasks on the \chembl~and \broad~datasets, respectively. This suggests that incorporating cell context into representation learning would be beneficial. That said, existing multi-modal baselines (InfoCORE, CLOOME) only outperform molecular structure-based approaches on \chembl and \toxcast, as they do not construct molecular representations holistically by using all cell-related modalities.

(3) \textbf{\method achieves the best average performance on all tasks compared to 27 baselines.} The improvements from \method range from 2.5\% to 6.4\% on average across four datasets compared to the second-best method. These gains are more significant when using the 80\% AUC threshold on classification datasets. While InfoCORE (GE) performs best among baselines on the \chembl~and \toxcast~datasets, it struggles to align molecular representations with more than two modalities and sometimes leads to negative transfer, as seen in \broad~and \biogen.

\vspace{-0.1in}
\subsection{RQ2: Molecule-Morphology Cross-Modal Matching}
\begin{table}[t]
\vspace{-0.1in}
\caption{Retrieval results on \chembl~(top) and \broad~(bottom): Left tables display ranking metrics for top candidates. Right figures visualize the distribution of rankings for the correct matching.}
\label{tab:matching-result}
\centering
\begin{minipage}{0.48\textwidth}
\centering
\begin{adjustbox}{width=0.92\linewidth}
\begin{tabular}{lcccc}
\toprule
\multirow{2}{*}{\chembl} & \multicolumn{2}{c}{NDCG \% ($\uparrow$)} & \multicolumn{2}{c}{HIT \% ($\uparrow$)} \\
\cmidrule{2-5}
& \multicolumn{1}{c}{top-1} & \multicolumn{1}{c}{top-10} & \multicolumn{1}{c}{top-1} & \multicolumn{1}{c}{top-10} \\
\midrule
CLOOME & 0 & 2.0 & 0 & 6.3 \\
InfoCORE & 0 & 4.5 & 0 & 11.3 \\
InfoAlign & \textbf{1.3} & \textbf{5.7} & \textbf{1.3} & \textbf{12.5} \\
\bottomrule
\end{tabular}
\end{adjustbox}
\end{minipage}
\begin{minipage}{0.5\textwidth}
\centering
\vspace{-0.1in}
\includegraphics[width=0.88\textwidth]{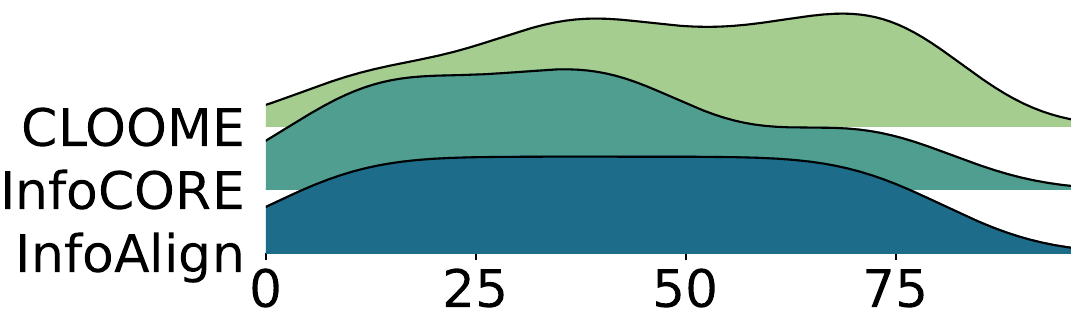}
\end{minipage}
% Second Table and its Corresponding Figure
\begin{minipage}{0.48\textwidth}
\centering
\begin{adjustbox}{width=0.95\linewidth}
\begin{tabular}{lrrrr}
\multirow{2}{*}{\broad} & \multicolumn{2}{c}{NDCG \% ($\uparrow$)} & \multicolumn{2}{c}{HIT \% ($\uparrow$)} \\
\cmidrule{2-5}
& \multicolumn{1}{c}{top-1} & \multicolumn{1}{c}{top-10} & \multicolumn{1}{c}{top-1} & \multicolumn{1}{c}{top-10} \\
% \toprule
% \broad \% ($\uparrow$) & NDCG@1 & NDCG@10 & HIT@1 & HIT@10 \\
\midrule
CLOOME & 0.5 & 0.9 & 0.5 & 1.5 \\
InfoCORE & \textbf{1.0} & \textbf{2.5} & \textbf{1.0} & 4.6 \\
InfoAlign & 0.5 & 2.3 & 0.5 & \textbf{5.1} \\
\bottomrule
\end{tabular}
\end{adjustbox}
\end{minipage}%
\begin{minipage}{0.5\textwidth}
\centering
\includegraphics[width=0.88\textwidth]{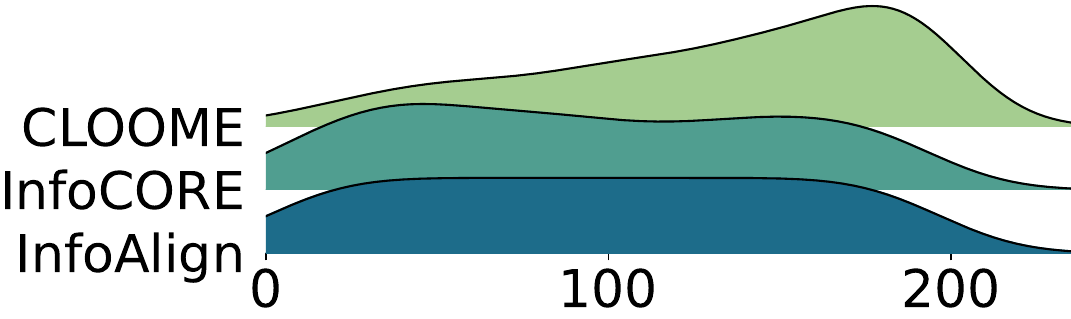}
\vspace{-0.1in}
\end{minipage}
\end{table}

\subsubsection{Experimental Setting}

We evaluated zero-shot matching performance of various methods for predicting cell morphology from query molecules, including baselines CLOOME and InfoCORE (CP) with pretrained encoders. For retrieval, we calculate the cosine similarity between the molecular representation and all cell morphology candidates, rank these candidates, and compute Normalized Discounted Cumulative Gain (NDCG) and HIT scores for the top-1 and top-10 candidates as metrics. To ensure a fair evaluation of zero-shot matching, we exclude the cell morphology data for molecules that were used to train the baseline encoders. Consequently, we have 80 molecule-cell morphology pairs from \chembl~and 196 pairs from \broad. All the morphology data are used as candidates for matching.

For \method, we use the pretrained decoder from~\cref{sec:pretrain-setup} to extract the morphology features of the encoded molecule and then calculate the likelihood of these decoded features against the candidate morphology data. We then rank the candidates in the decoding space based on their likelihood scores.

\subsubsection{Results and Analysis}
Cross-modal matching results are in~\cref{tab:matching-result}. \method outperforms InfoCORE on \chembl and is comparable to InfoCORE on \broad, with both surpassing CLOOME. Additionally, we visualized the distribution of ranking positions for correct matching pairs to compare overall retrieval performance. The results show that \method and InfoCORE perform similarly, while CLOOME consistently ranks correct pairs lower.

\subsection{RQ3: Performance Analysis}
\subsubsection{Ablation Studies}
\begin{table*}[t]
    \caption{Ablation studies on the pretraining loss. Different node features are removed from the context graph to assess their impact on downstream tasks. Avg. AUC is reported.}
    \label{tab:ablation-loss}
    \renewcommand{\arraystretch}{0.9}
    \renewcommand{\tabcolsep}{0.8mm}
    \centering
    \begin{adjustbox}{width=0.7\linewidth}
    \begin{tabular}{lcccc}
    \toprule
     & \shortstack{ChEMBL2K \\ AUC $\uparrow$} & \shortstack{Broad6K \\ AUC $\uparrow$} & \shortstack{ToxCast \\ AUC $\uparrow$} & \shortstack{Biogen3K \\ MAE $\downarrow$ ($\times$100)} \\
    \midrule
    % Best Baseline      & $79.3 \pm 0.9$ & $67.3 \pm 0.4$ & $65.3 \pm 0.1$ & $52.8 \pm 0.2$ \\
    Default as~\cref{eq:loss}      & $81.3 \pm 0.6$ & $70.0 \pm 0.1$ & $66.4 \pm 1.1$ & $49.4 \pm 0.2$ \\
    w/o Cell Morphology         & $80.7 \pm 0.6$ & $68.6 \pm 0.1$ & $65.5 \pm 1.1$ & $51.7 \pm 1.1$ \\
    w/o Gene Expressions        & $78.3 \pm 0.5$ & $68.6 \pm 0.2$ & $64.7 \pm 1.0$ & $50.3 \pm 0.5$ \\
    w/o Molecular Features      & $79.1 \pm 0.2$ & $67.1 \pm 0.4$ & $65.8 \pm 2.3$ & $51.7 \pm 0.6$ \\
    \bottomrule
    \end{tabular}
    \end{adjustbox}
\end{table*}

We perform ablation studies on~\cref{eq:loss} by pretraining encoders with different targets removed: (1) molecule-related, (2) cell morphology-related, and (3) gene expression-related features. 
The results in~\cref{tab:ablation-loss} cover all datasets. We observe that both cell morphology and gene expression features are crucial for achieving the best performance. Different biological targets have varying impacts across datasets: molecular structure has more influence on \broad and \biogen, while gene expression is more important for \chembl and \toxcast.

\begin{figure}[t]
    \centering
    \begin{subfigure}[b]{0.49\textwidth}
        \centering
        \includegraphics[width=0.85\linewidth]{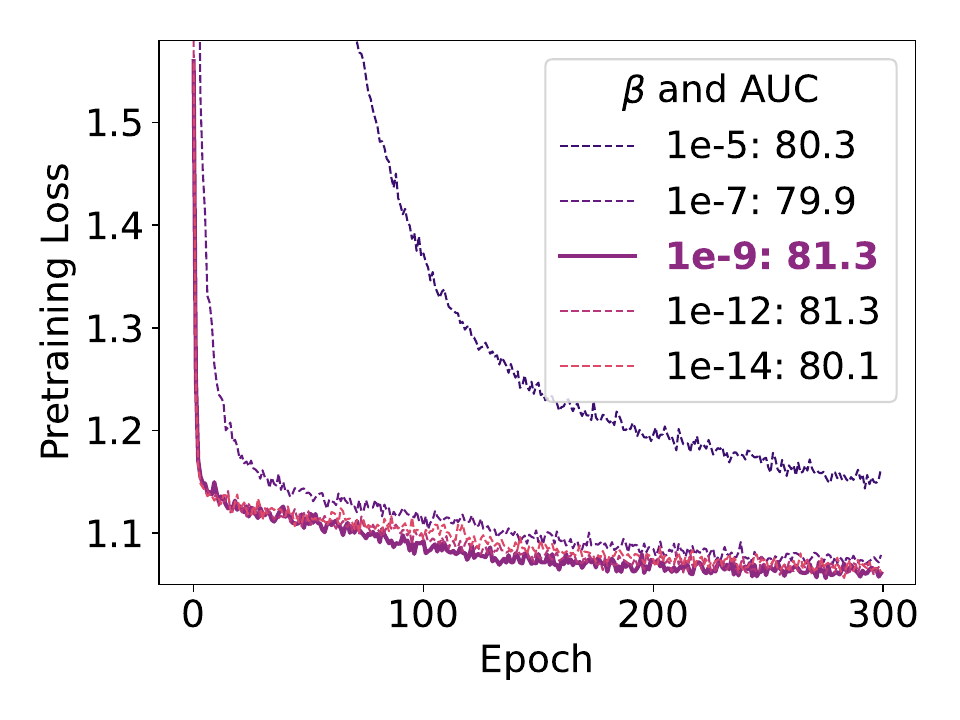}
        % \vspace{-0.2in}
        \caption{\small Losses and AUC on varying \(\beta\) (Default: 1e-9).}
        \label{fig:loss-prior}
    \end{subfigure}%
    \begin{subfigure}[b]{0.49\textwidth}
        \centering
        \includegraphics[width=0.85\linewidth]{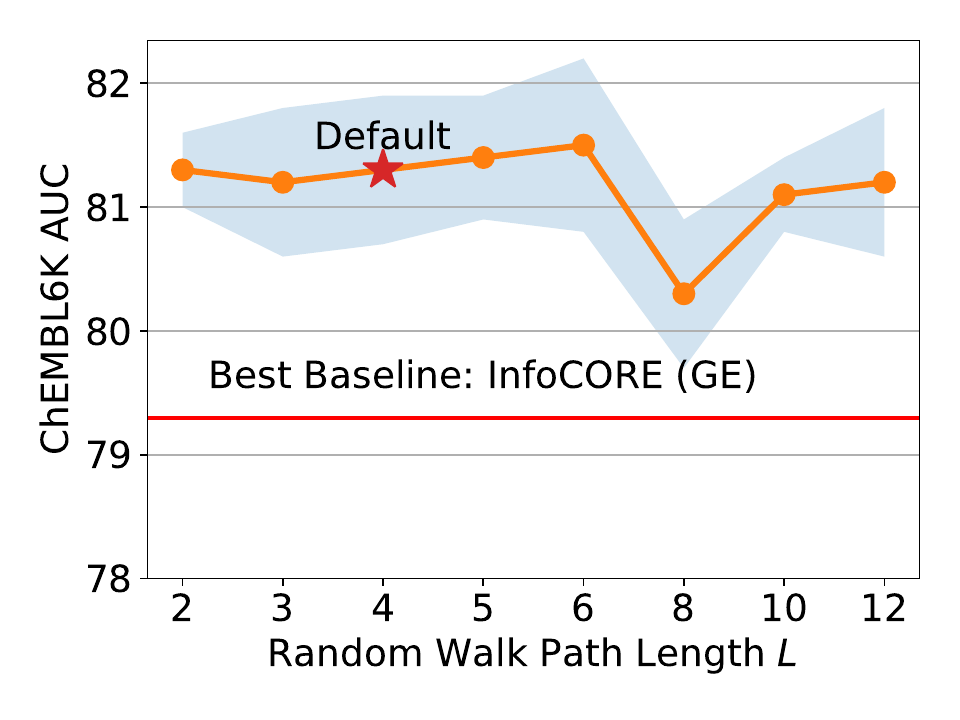}
        % \vspace{-0.2in}
        \caption{AUC on varying $L$ (Default: 4).}
        \label{fig:loss-walk}
    \end{subfigure}
    \caption{Analysis on the hyperparameters: strength of prior \(\beta\) and random walk length \(L\). AUC is computed on the test set of \chembl. }
    \label{fig:loss-ablation}
    \vspace{-0.2in}
\end{figure}
\subsubsection{Hyperparameter Analysis}
Lastly, we perform analysis for the hyperparameters: the strength of the regularization to the prior Gaussian distribution $\beta$ and the length of the random walk paths $L$. Results are presented in~\cref{fig:loss-ablation}. We observe a trade-off between the principles of minimality and sufficiency in \cref{fig:loss-prior}: a too-high $\beta$ value (minimal information) makes it challenging for the representation to be sufficiently expressive for molecular, gene expression, and cell morphology features, potentially degrading downstream performance. Conversely, a too-low $\beta$ value weakens minimality and may impair generalization. The convergence of the pretraining loss could serve as a good indicator to balance these aspects. For the hyperparameter $L$, we observe in \cref{fig:loss-walk} that downstream performance on \chembl is relatively robust across a wide range of walk lengths. 

\section{Conclusion}
\label{sec:conclusion}
In this work, we proposed learning molecular representations in a cell context with three modalities: molecular structure, gene expression, and cell morphology. We introduced the information bottleneck approach, \method, using a molecular graph encoder and multiple MLP decoders. \method learned minimal sufficient molecular representations extracted by reconstructing features in the random walk path on a cellular context graph. This context graph incorporated molecules, cell morphology, and gene expression information defined in scalar or vector spaces to construct nodes, and used various chemical, biological, and computational criteria to define their weighted edges. We demonstrated the theoretical and empirical advantages of the proposed method. \method outperformed other representation learning methods in various molecular property prediction and zero-shot molecule-morphology matching tasks.

\subsubsection*{Acknowledgments}
This study was supported by National Institutes of Health (R35 GM122547 to AEC) and an internship funded by the Massachusetts Life Sciences Center (to GL).

\bibliography{reference}
\bibliographystyle{plainnat}

%%%%%%%%%%%%%%%%%%%%%%%%%%%%%%%%%%%%%%%%%%%%%%%%%%%%%%%%%%%%

\appendix
\onecolumn
\newpage \section{Proof Details}\label{sec:more-theory}

\subsection{Proof of \cref{eq:mi-lower-upper-bo}}\label{sec:proof-lower-upper}

For the input, latent, and target variables $X$, $Z$, and $Y$, the exact computation of the mutual information (MI) $I(Z; Y)$ and $I(X; Z)$ is intractable. For the molecule $x$, its latent representation $z$, and any biological target from cellular responses $y$, we introduce variational approximations $q(y | z)$ to obtain a lower bound on $I(Z; Y)$:
\begin{equation}
\begin{aligned}
    I(Z; Y)  
    & = \mathbb{E}_{p(z, y)}\left[ \log \frac{p(z, y) q(y | z)}{p(y)p(z)q(y | z)}\right], \\
    & = \mathbb{E}_{p(z, y)}\left[ \log \frac{q(y | z)}{p(y)}\right] + \mathbb{E}_{p(z)}\left[ \operatorname{KL}\left(p(y | z) \| q(y | z)\right) \right], \\
    & \geq \mathbb{E}_{p(z, y)} \left[ \log q(y | z)\right] + H(Y) \triangleq I_{DLB} \\
\end{aligned}
\end{equation}
This is because that $\operatorname{KL}\left(p(y | z) \| q(y | z)\right) \geq 0$.
We introduce the variational approximations $q(z)$ for a upper bound on $I(X; Z)$:
\begin{equation}
\begin{aligned}
    I(X; Z) 
    & = \mathbb{E}_{p(x, z)}\left[ \log \frac{p(x, z) q(z)}{p(x)p(z)q(z)}\right], \\
    & = \mathbb{E}_{p(x, z)}\left[ \log \frac{p(z|x)}{q(z)}\right] - \mathbb{E}_{p(z)} \left[(p(z) \| q(z) \right], \\
    & \leq \mathbb{E}_{p(x)} \left[\operatorname{KL}\left( p(z|x) \| q(z) \right) \right] \triangleq I_{EUB}
\end{aligned}
\end{equation}

\subsection{Proof of \cref{propos:bound-diff}}\label{sec:proof-bound-diff}

For the molecule $x$, its latent representation $z$, and any biological target from cellular responses $y$, we use the neural network parameterized critic $h(z, y)$ with the energy-based variational family for density approximation~\citep{poole2019variational}:
$$
q(y | z) = \frac{p(y)}{\mathbb{E}_{p(y)}\left[ e^{h(z, y)}\right]} e^{h(z,y)}.
$$
Thus, we can rewrite $I_{DLB}$ based on the unnormalized distribution of $q(y | z)$:
\begin{equation}\label{eq:ulb}
\begin{aligned}
    I_{DLB} & = \mathbb{E}_{p(z, y)} \left[ \log q(y | z)\right] + H(Y) \\
    & = \mathbb{E}_{p(z, y)}\left[ \log \left( \frac{p(y)}{\mathbb{E}_{p(y)} \left[e^{h(z, y)}\right]} e^{h(z,y)} \right) \right] - \mathbb{E}_{p(y)} \left[ \log p(y)\right], \\
    & = \mathbb{E}_{p(z, y)}\left[h(z, y)\right] - \mathbb{E}_{p(z, y)}\left[ \mathbb{E}_{p(y)}[ e^{h(z, y)}] \right] , \\
    & = \mathbb{E}_{p(z, y)}\left[h(z, y)\right] - \mathbb{E}_{p(z)} (\log \Zp(z)),
\end{aligned}
\end{equation}
where $\Zp(z) = \mathbb{E}_{p(y)}[ e^{h(z, y)}]$ is the partition function.

Note that the log partition function is intractable. \citet{poole2019variational} introduced a new variational parameter $a(\cdot)$ to upper bound $\Zp(z)$, deriving a tractable lower bound for $I_{DLB}$:
\begin{equation}
\begin{aligned}
    I_{DLB} \geq \mathbb{E}_{p(z, y)} [h(z, y)] - \mathbb{E}_{p(z)}\Bigg[ \frac{\mathbb{E}_{p(y)} [e^{h(z,y)}]}{a(z)} \quad + \log(a(z)) - 1 \Bigg].
\end{aligned}
\end{equation}
This is because $\forall x, a >0$, the inequality $\log(x) \leq \frac{x}{a} + \log(a) - 1$ holds, which can be applied to the second term of~\cref{eq:ulb}. The $I_{NWJ}$ bound~\citep{nguyen2010estimating} is a special case where $a(z) = e$.
\begin{equation}
\begin{aligned}
    I_{NWJ} & \triangleq \mathbb{E}_{p(z, y)} [h(z, y)] - \mathbb{E}_{p(z)}\Bigg[ \frac{\mathbb{E}_{p(y)} [e^{h(z,y)}]}{e} \quad + \log(e) - 1 \Bigg] \\
    & = \mathbb{E}_{p(z, y)} [h(z, y)] - e^{-1} \mathbb{E}_{p(z)} [\Zp(z)].
\end{aligned}
\end{equation}
$I_{NWJ}$ has high variance due to the estimation of the upper bound on the log partition function. Based on $I_{NWJ}$ and multiple examples, one can derive the encoder-based lower bound $I_{ELB}$ for InfoNCE. 

Suppose there are $K-1$ additional examples independently and identically sampled and denoted as $y_{2:K}$, and the critic is configured with parameters $a(\cdot)$ as $1 + \log \frac{e^{h(z,y)}}{a(z; y, y_{2:K})}$. Then, we can rewrite $I_{NWJ}$ for its multi-sample version:
\begin{equation}\label{eq:nwj-multi-sample}
\begin{aligned}
    I_{NWJ}
    & = \mathbb{E}_{p(z, y)p(y_{2:K})} \left[ 1 + \log \frac{e^{h(z, y)}}{a(z; y, y_{2:K})} \right] - e^{-1} \mathbb{E}_{p(y)p(z)p(y_{2:K})} \left[ e^{1 + \log \frac{e^{h(z, y)}}{a(z; y, y_{2:K})} } \right], \\
    & = 1 + \mathbb{E}_{p(z, y)p(y_{2:K})} \left[ \log \frac{e^{h(z, y)}}{a(z; y, y_{2:K})} \right] - \mathbb{E}_{p(y)p(y_{2:K})p(z)} \left[ \frac{e^{h(z, y)}}{a(z; y, y_{2:K})} \right].
\end{aligned}
\end{equation}

Multiple samples can be utilized for the Monte Carlo method $m(z; y, y_{2:K})$ to estimate the upper bound on the partition function $a(z; y, y_{2:K})$:
$$
a(z; y, y_{2:K}) = m(z; y, y_{2:K}) = \frac{1}{K} \left( e^{h(z, y)} + \sum_{i=2}^K e^{h(z, y_i)} \right),
$$
where $K-1$ independent samples are drawn from $\prod_{i}p({y_i})$ and one sample from $p(z, y)$ for the term $\mathbb{E}_{p(z, y)p(y_{2:K})}[\cdot]$ or $K$ samples from $\prod_{i=1}^K p(y_i)$ (we set $y_1=y$) for a $p(z)$ sample in the $\mathbb{E}_{p(y)p(y_{2:K})}[\cdot]$ term.
Therefore, we can derive $I_{ELB} \triangleq I_{NCE}$:
\begin{equation}\label{eq:elb-add}
\begin{aligned}
    I_{ELB} \triangleq I_{NCE} 
    & = 1 + \mathbb{E}_{p(z, y)p(y_{2:K})} \left[ \log \frac{e^{h(z, y)}}{m(z; y, y_{2:K})} \right] - \mathbb{E}_{p(y)p(y_{2:K})p(z)} \left[ \frac{e^{h(z, y)}}{m(z; y, y_{2:K})} \right], \\
    & = 1 + \mathbb{E}_{p(Y | Z)p(z)p(y_{2:K})} \left[ \log \frac{e^{h(z, y)}}{\frac{1}{K}\sum_{i=1}^K e^{h(z, y_i)}} \right] \\ & \quad\quad - \mathbb{E}_{p(y)p(y_{2:K})p(z)} \left[ \frac{e^{h(z, y)}}{\frac{1}{K}\sum_{i=1}^K e^{h(z, y_i)}} \right], \\
    & = \mathbb{E}_{p(z, y)} \left[ h(z, y) \right] 
    -  \mathbb{E}_{p(z)} \left[\log \frac{1}{K}\sum_{i=1}^K e^{h(z, y_i)}\right].
\end{aligned}
\end{equation}
Note that for $\mathbb{E}_{p(y)p(y_{2:K})p(z)}[\cdot]$, we average the bound over $K$ replicates as well to ensure that the last term in~\cref{eq:nwj-multi-sample} is the constant 1. Now, $I_{ELB}$ or $I_{NCE}$ is upper bounded by $\log K$, rather than $a(\cdot)$.
Hence, the difference between $I_{DLB}$ and $I_{ELB}$ is
\begin{equation}\label{eq:diff-dlb-elb}
\begin{aligned}
    I_{DLB} - I_{ELB}
    & = \mathbb{E}_{p(z)} \left[\log \frac{1}{K}\sum_{i=1}^K e^{h(z, y_i)}\right] -  \mathbb{E}_{p(z)} (\log \Zp(z)) \geq 0.
\end{aligned}
\end{equation}
When $K$ is sufficiently large to estimate the partition function, we have $\mathbb{E}_{p(z)} \left[ \log \left( \mathbb{E}_{p(y)} \left[e^{h(z, y)}\right] \right) \right]$ for the left term, indicating that $I_{DLB} - I_{ELB} = 0$. Since $I_{NCE}$ is upper bounded by $\log K$~\citep{oord2018representation}, smaller values of $K$ may result in a less tight $I_{ELB}$, causing $I_{DLB} - I_{ELB} \geq 0$ to always hold. In particular, $I(Z; Y)  > \log K$ implies that the bound $I_{ELB}$ will be loose.

\section{Context Graph Details}\label{sec:more-context-graph}

\begin{figure}[t]
    \centering
    \includegraphics[width=1.\linewidth]{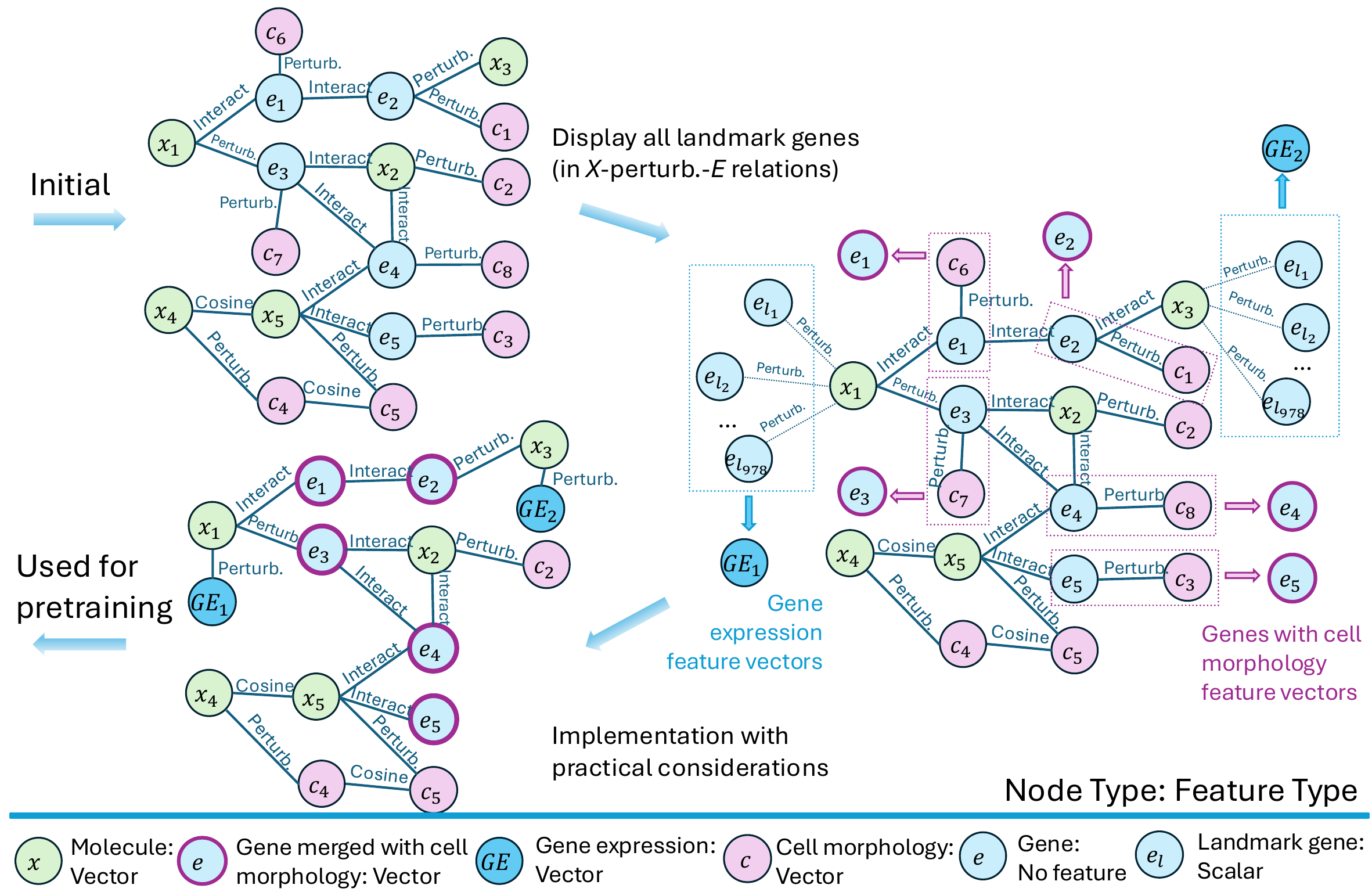}
    \caption{From the initial idea in~\cref{sec:method} to the practical implementation of the context graph, we first display relations between molecules and all the landmark genes from~\cite{wang2016drug} for the $X_1 - E_3$ and $X_3 - E_2$ relationships. $E_3$ and $E_2$ are landmark genes involved in small molecule perturbations and cell morphology perturbation; we display them separately for clarity. Next, we merge all landmark genes into new gene expression nodes and integrate genes from genetic perturbations in the JUMP dataset~\citep{chandrasekaran2023jump} with cell morphology features. Practical considerations are detailed in~\cref{sec:pretrain-setup,sec:more-context-graph}.}
    \label{fig:path-details}
\end{figure}

\subsection{Edge Construction}\label{sec:more-edge-construct}
Edges represent similarity relationships between molecules, genes and cells. According to the chemical or biological criteria, we have following types of edges:
\begin{enumerate}
    \item Molecule-Cell Morphology Edges: These edges are introduced through molecule perturbation experiments from cell painting datasets created by~\cite{bray2016cell} and the JUMP dataset~\citep{chandrasekaran2023jump}. It links molecule nodes with cell morphology nodes. We use the edge weight $1$ for all these edges.
    \item Edge-Cell Morphology Edges: These edges should be introduced by genetic perturbation from the JUMP dataset~\cite{chandrasekaran2023jump}. The perturbations are either based on gene overexpression (ORF) or gene knockout techniques (CRISPR). They link the gene nodes and the cell morphology nodes. However, the genes introduced by the genetic perturbations lack gene expression profiling from~\citep{subramanian2017next} as node features. We did not implement gene-cell morphology edges from \citep{chandrasekaran2023jump} due to the absence of differential gene expression profiling values \citep{subramanian2017next}. Instead, we merged the gene nodes from \citep{chandrasekaran2023jump} with their linked cell morphology nodes, creating single nodes. This approach enables a more efficient context graph, incorporating some gene nodes with cell morphology features.
    \item Molecule-Gene Edges: These edges could represent molecule-gene binding and regulation relationships, linking molecules to genes~\citep{himmelstein2017systematic}. Some links can be sourced from~\citep{himmelstein2017systematic}, and we also retrieve gene-molecule links from~\citep{wang2016drug} by selecting the top 5\% absolute differential expression values.
    \item Gene-Gene Edges: These edges denote the relationships of gene-gene covariance and interaction and we use the links from~\citep{himmelstein2017systematic}.
\end{enumerate}

We enrich the edges in the context graph by incorporating computational similarity edges, where cosine similarity is computed among within nodes having the same type and feature vectors. We note that the cell morphology features from~\citep{bray2016cell} and~\citep{chandrasekaran2023jump} have different dimensions since the latter has applied batch correction techniques~\cite{arevalo2023evaluating} on the CellProfiler features~\citep{carpenter2006cellprofiler}. Thus, we cannot compute the similarity between these two subsets of cell morphology nodes.
We use (1) a 0.8 similarity threshold and (2) a minimal sparsity of 99.5\% by selecting top 0.5\% similar edges to avoid excessive noise in computational similar edges.

\subsection{Dataset Sources of Nodes}\label{sec:more-node-source}
Here are the datasets we used to create different types of nodes on the context graph:
\begin{itemize}
    \item Molecule nodes: Molecular nodes are sourced from two cell painting datasets: one by~\citet{bray2017dataset} and the other from the recently released JUMP dataset~\citep{chandrasekaran2023jump}, and the third source from~\citet{wang2016drug}, which are used to study adverse drug reactions.
    \item Gene nodes: Gene nodes are from the landmark genes used by~\citet{wang2016drug} in creating the LINCS L1000 profiling of drugs. Other gene nodes come from genetic perturbations in the JUMP dataset~\citep{chandrasekaran2023jump}. The gene nodes from~\citep{chandrasekaran2023jump} have cell morphology features as described in~\cref{sec:more-edge-construct}. The landmark gene nodes from~\citep{wang2016drug} have scalar gene expression profiles, but these values are updated in the new gene expression nodes.
    \item Cell morphology nodes: Cell nodes are sourced from the two cell painting datasets~\citep{bray2016cell,chandrasekaran2023jump}.
    \item Gene expression nodes: Based on landmark genes from~\citep{wang2016drug}, each gene expression node summarizes all gene expression profiles into vectors from a small molecule perturbation. Since~\citet{wang2016drug} measured the same landmark genes for a set of molecules, we update new gene expression nodes with feature vectors for all these landmark genes. This approach efficiently constructs decoding targets from molecules to gene expression profiles and prevents redundant gene-molecule connections.
\end{itemize}

We present an example of the cellular context graph in~\cref{fig:path-details}.

\section{Experiment Details}\label{sec:more-experiment}

\begin{figure}[t]
    \centering
    \includegraphics[width=0.7\linewidth]{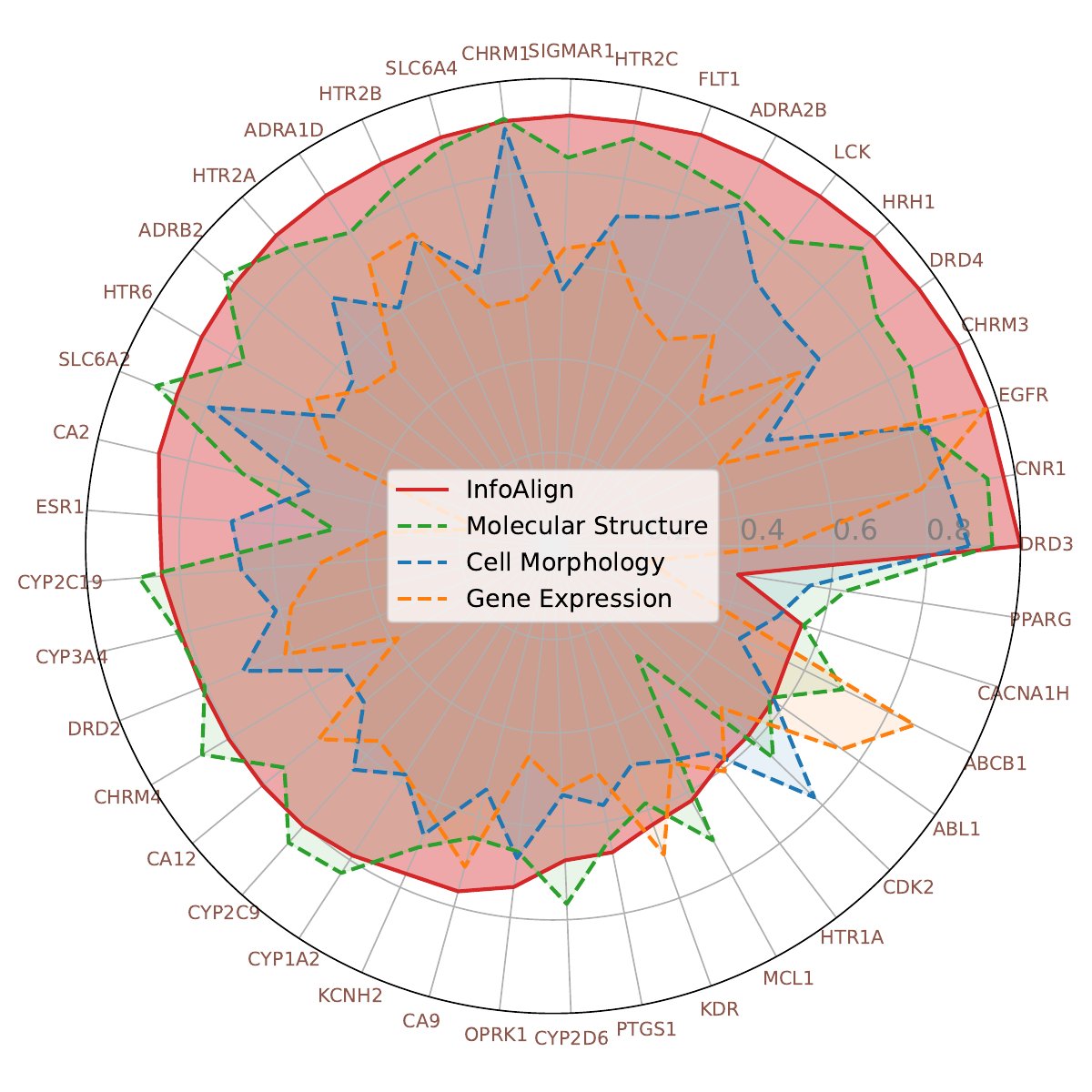}
    \caption{An overview of the representation's predictive performance on all 41 bioactivity prediction tasks in \chembl. Results for molecular structure are obtained from the best method ContextPred. Results for cell morphology and gene expression come from the best method based on MLPs.}
    \label{fig:complete-chembl}
\end{figure}

\begin{figure}[t]
    \centering
    \includegraphics[width=0.5\linewidth]{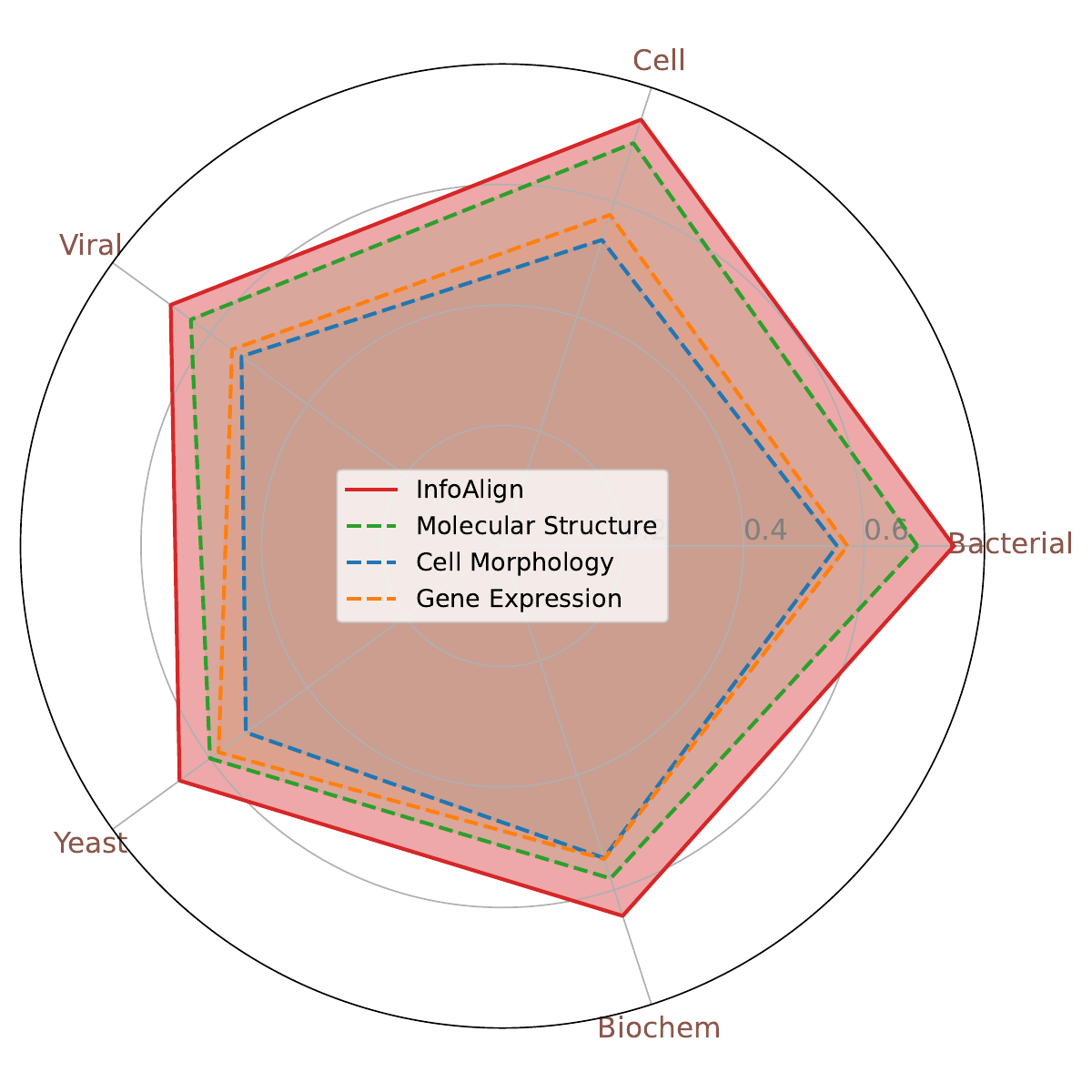}
    \caption{An overview of the representation's predictive performance across five major task categories on \broad. Results for molecular structure are obtained from the best method based on JOAO. Results for cell morphology come from the best method based on RF. Results for gene expression are derived from the best method based on MLP.}
    \label{fig:complete-broad}
\end{figure}

\subsection{Prediction Datasets}\label{sec:more-exp-setup}
All experiments were run on a single 32G V100.
Prediction dataset statistics are in~\cref{tab:task-data-info}:
\begin{itemize}
    \item \chembl~\citep{gaulton2012chembl}: The dataset is a subset of the ChEMBL dataset~\citep{gaulton2012chembl}, overlapping with the JUMP CP~\citep{chandrasekaran2023jump} datasets. We determined activity using the ``activity\_comment'' provided by ChEMBL. If not, we applied a threshold of 6.5, labeling compounds with pChEMBL > 6.5 as active. We exclude all molecules in the dataset from the pretraining set to avoid data leakage. There are a total of 41 tasks related to protein binding affinity, which are converted to binary activity values. We filter the dataset to ensure that each task has at least one positive and five negative examples. 
    \item \broad~\citep{moshkov2023predicting}: The original version provided by~\citet{moshkov2023predicting} is a collection of 16,170 molecules tested in 270 assays, resulting in a total of 585,439 readouts. However, there are a large number of missing values, with 153 assays having a missing value percentage above 99\%. To mitigate bias in the conclusions, we extract subsets where the percentage is less than 50\%.
    \item \toxcast~\citep{richard2016toxcast}: The toxicology data is collected from the ``Toxicology in the 21st Century'' initiative, widely utilized in many graph machine learning models~\citep{hu2020open}. The dataset comprises 8,576 molecules and 617 binary classification tasks.
    \item \biogen~\citep{fang2023prospective}: The dataset includes properties that describe the disposition of a drug in the body, including absorption, distribution, metabolism, and excretion (ADME). It is collected from 120 Biogen datasets across six ADME in vitro endpoints over 20 time points spanning about 2 years. The endpoints include human liver microsomal (HLM) stability reported as intrinsic clearance (Clint, mL/min/kg), MDR1-MDCK efflux ratio (ER), Solubility at pH 6.8 (µg/mL), rat liver microsomal (RLM) stability reported as intrinsic clearance (Clint, mL/min/kg), human plasma protein binding (hPPB) percent unbound, and rat plasma protein binding (rPPB) percent unbound.
\end{itemize}

We utilize scaffold-splitting with a ratio of 0.6:0.15:0.25 and follow~\citep{hu2020open} for the \toxcast~dataset. 
We use the Area under the curve (AUC) score for classification and mean absolute error (MAE) for regression.
We report the mean and standard deviations from ten runs.

\subsection{Implementation and Baseline}\label{sec:more-implement-baseline}
We consider baselines from three representation sources: molecular structures, cell morphology, and gene expressions. Moreover, we have three different ways to represent molecular structures, including fingerprints based on domain knowledge, GNNs based on the graph structure of molecules, and chemical language models (ChemLM) based on SMILES-sequence structure of molecules.

\begin{enumerate}
    \item Molecular descriptors/fingerprints~\citep{rogers2010extended} (Structure only): We train MLPs, Random Forests (RF), and Gaussian Processes (GP) on these representations.
    \item Pretrained GNN representations~\citep{hu2020strategies} (Structure only): We consider AttrMask, ContextPred, and EdgePred with supervised pretraining~\citep{hu2020strategies}. We also include GraphCL~\citep{you2020graph}, GROVER~\citep{rong2020self}, JOAO~\citep{you2021graph}, MGSSL~\citep{zhang2021motif}, GraphLoG~\citep{xu2021self}, GraphMAE~\citep{hou2022graphmae}, DSLA~\citep{kim2022graph}, and UniMol~\citep{zhou2023unimol}. We implement GraphCL, GROVER, and JOAO based on~\citep{wang2024evaluating}. Fine-tuned MLPs are applied on top of the pretrained representations.
    \item Pretrained ChemLM representations~\citep{frey2023neural} (Structure only): We consider pretrained models such as 102M Roberta and 87M GPT2 implemented by~\citep{datamol2024}. We also include MolT5~\citep{edwards2022translation} and 19M ChemGPT~\citep{frey2023neural}. We apply fine-tuned MLPs on top of these pretrained representations.
    \item Cell Morphology~\citep{rogers2010extended} (Cell or Structure only): Cell morphology features are available in for part of molecules in the~\chembl~and~\broad~datasets. We train MLPs, RF, and GP on these representations. Note that not all molecules have corresponding cell morphology feature vectors; in such cases, we replace the predictions on the missing feature with ML predictions on the structure.
    \item Gene Expression~\citep{rogers2010extended} (Gene or Structure only): Differential gene expression values are available for part of molecules in the~\chembl~and~\broad~datasets. We train MLPs, RF, and GP on these representations. Note that not all molecules have corresponding gene expression vectors over landmark genes; in such cases, we replace the predictions on the missing feature with ML predictions on the structure.
    \item CLOOME~\citep{sanchez2023cloome} and InfoCORE~\citep{wang2023removing} (Structure-Cell or Structure-Gene aligned): CLOOME utilizes ResNet~\citep{he2016deep} and descriptor-based MLP to align representation from cell morphology images with the molecular structure representation. We use their pretrained MLP to obtain molecular representations and fine-tune another MLP on top of these representations. InfoCORE has two versions, InfoCORE-CP and InfoCORE-GE, which align the molecular graph representation with cell morphology features or differential gene expression features, respectively. We use both versions as baselines and fine-tune another MLP on top of these representations.
\end{enumerate}

% Results for the \toxcast and \biogen datasets are provided in \cref{tab:complete-toxcast} and \cref{tab:complete-biogen}, respectively.
\subsection{More Results for Molecular Property Prediction}\label{sec:more-assay-result}

We present additional comparisons on the \chembl dataset between basic representation approaches and \method across all task dimensions in \cref{fig:complete-chembl}. Similarly, results for the \broad dataset, comparing basic representations across five major task dimensions (Cell, Yeast, Viral, Biochem, and Bacterial-related targets), are shown in \cref{fig:complete-broad}. Combined with~\cref{tab:chembl-broad-ext,tab:tox-biogen-ext}, these detailed results lead to further observations:

(1) Different structure-based molecular representations vary in sensitivity to model architecture. Dramatic performance drops occur with Morgan FP when replacing the MLP architecture with RF or GP in the \chembl and \broad datasets. Conversely, in the \biogen dataset, RF and GP significantly outperform MLP. In contrast, pretrained GNN and ChemLM representations maintain more consistent performance across various datasets.

(2) Learning universal molecular representations solely from molecular structures remains challenging, even within the representation category. For pretrained GNN representations, ContextPred outperforms others on the \chembl dataset. JOAO excels on the \broad dataset. UniMol and GraphMAE are the best pretrained GNN representations on \toxcast and \biogen datasets, respectively. For ChemLM representations, MolT5 excels over other sequential-based models in the \toxcast and \biogen datasets, but this is not the case with the \chembl and \broad datasets. Different datasets may emphasize varied aspects of bioactivity classification or regression and pose generalization challenges for molecular representation learning.

(3) \method shows strong generalization for the targets of non-human cells, as shown in~\cref{fig:complete-broad}.
Although the context graph primarily uses data from small molecule and genetic perturbation datasets~\citep{bray2016cell,chandrasekaran2023jump} focused on human cell cultures, \method also exhibits robust generalization to bacterial and viral targets compared to basic representation approaches.

\end{document}